\PassOptionsToPackage{table,dvipsnames}{xcolor}
\documentclass{sjtudenglab}

\usepackage{amsmath}
\usepackage{enumerate}
\usepackage{amsfonts}
\usepackage{amsthm}
\usepackage{newtxtt}
\usepackage{diagbox}
\usepackage{colortbl}
\usepackage{amssymb}
\usepackage{xspace}
\usepackage{wrapfig}
\usepackage{adjustbox}
\usepackage{array}
\usepackage{tabularx}
\usepackage{booktabs}
\usepackage{mathtools}
\usepackage{tikz}
\usepackage{enumitem}
\usepackage{silence}
\usepackage{dsfont}
\usepackage[table]{xcolor}
\usepackage[dvipsnames]{xcolor}
\usepackage{multirow}
\usepackage{makecell}
\usepackage{xfakebold}
\usepackage{siunitx}
\usepackage{thmtools}

\usepackage{amsmath,amsfonts,bm}

\def\eqref#1{equation~\ref{#1}}

\def\1{\bm{1}}

\DeclareMathAlphabet{\mathsfit}{\encodingdefault}{\sfdefault}{m}{sl}
\SetMathAlphabet{\mathsfit}{bold}{\encodingdefault}{\sfdefault}{bx}{n}

\usepackage{parskip}

\usepackage{natbib}
\usepackage{amsfonts,bm}

\usepackage{amsmath, amssymb,mathrsfs, amsthm, mathtools}
\usepackage{xspace}
\usepackage{enumitem}
\usepackage{lipsum}
\usepackage{xpatch}
\usepackage{mathtools}
\usepackage{dsfont}
\usepackage{pgf,tikz}
\usetikzlibrary{positioning,matrix,patterns}
\usepackage{caption}
\usepackage{graphicx}
\usepackage{makecell}
\usepackage{multirow}
\usepackage[mathscr]{euscript}
\definecolor{hanblue}{rgb}{0.27, 0.42, 0.81}
\definecolor{deepred}{HTML}{900C3F}
\definecolor{deepgreen}{HTML}{2F6960}
\hypersetup{
    colorlinks=true,
    linkcolor=hanblue,
    urlcolor=hanblue,
    citecolor=hanblue,
    anchorcolor=hanblue}

\declaretheoremstyle[
  headfont=\sffamily\bfseries,
]{sansserif}

\theoremstyle{sansserif}

\theoremstyle{definition}

\theoremstyle{sansserif}

\theoremstyle{remark}

\DeclarePairedDelimiter\abs{\lvert}{\rvert}%
\DeclarePairedDelimiter\norm{\lVert}{\rVert}%

\makeatletter
\let\oldabs\abs
\def\abs{\@ifstar{\oldabs}{\oldabs*}}
\let\oldnorm\norm
\def\norm{\@ifstar{\oldnorm}{\oldnorm*}}
\makeatother

\definecolor{textgray}{HTML}{6E6E73}
\usetikzlibrary{positioning, calc}
\usetikzlibrary{decorations.pathmorphing}

\makeatletter
\patchcmd{\wrong@fontshape}{\@gobbletwo}{}{}{}
\makeatother
\numberwithin{equation}{section}
\makeatletter
\AtBeginDocument{
  \urlstyle{sf}
  
}
\makeatother

\renewcommand{\eqref}[1]{\textup{(\ref{#1})}}
\setcitestyle{numbers,square,comma}

\renewcommand{\title}[1]{\newcommand{\titlelist}{{\sffamily\bfseries\fontsize{20}{24}\selectfont #1}}}
\newenvironment{acks}{\section*{Acknowledgments}}{}
\title{MirrorPPR: Exemplar-Based Portrait Photo Retouching}

\author[1]{Zhihong Liu}
\author[1]{Zheng Li}
\author[1]{Jiachun Jin}
\author[1,2]{Siqi Kou}
\author[1,2]{Yitao Jian}
\author[2]{Fengpei Yu}
\author[1]{Zhijie Deng}

\affiliation[1]{Shanghai Jiao Tong University}
\affiliation[2]{Triverse AI}

\abstract{

While text-guided image editing has made remarkable progress, it remains limited in structural portrait retouching. Textual descriptions struggle to convey fine-grained changes to facial features and body proportions.
To address this gap, we introduce Exemplar-Based Portrait Photo Retouching, where the model is given an exemplar pair and tasked with inferring and applying the same retouching operations to a new query image.
Existing exemplar-based editing methods primarily focus on tasks with pronounced visual transformations. In contrast, structural portrait retouching involves extremely delicate and localized modifications, making accurate extraction and transfer of these edits challenging.
To tackle this, we propose MirrorPPR, a novel framework designed to capture and transfer subtle structural retouching operations. Our method uses a Retouching Operation Extractor to capture the subtle differences from the exemplar pair. The extracted representations are then injected into a pre-trained Diffusion Transformer (DiT) through a connector and Low-Rank Adaptation (LoRA) modules.
Furthermore, constructing perfectly aligned cross-identity training pairs is severely hindered by operation misalignment. To overcome this, we propose an advanced data self-augmentation paradigm that ensures strictly aligned retouching operations.
To alleviate data scarcity and support this novel task, we introduce MirrorPPR47M, a large-scale dataset with over 47 million retouched pairs. By structuring the dataset into simulated and professional subsets, we enable progressive curriculum learning to smoothly optimize the network.
Extensive experiments demonstrate that MirrorPPR significantly outperforms existing baselines in both retouching quality and identity preservation.

}

\metadata[Correspondence]{\sffamily Zhijie Deng: \texttt{zhijied@sjtu.edu.cn}}
\date{\sffamily\today}
\metadata[Project Page]{\sffamily \url{https://sjtu-deng-lab.github.io/MirrorPPR}}

\AtBeginDocument{ 
  \fancyheadoffset[R]{-0.3cm}
  \fancyhead[R]{\raisebox{9pt}{\includegraphics[height=35pt]{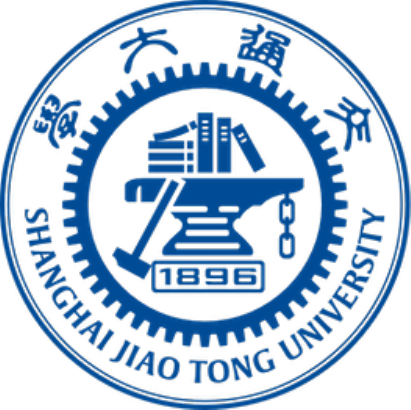}}}
}

\begin{document}

\maketitle

\section{Introduction}
\label{sec:intro}

\begin{figure}
    \centering
    \includegraphics[width=\linewidth]{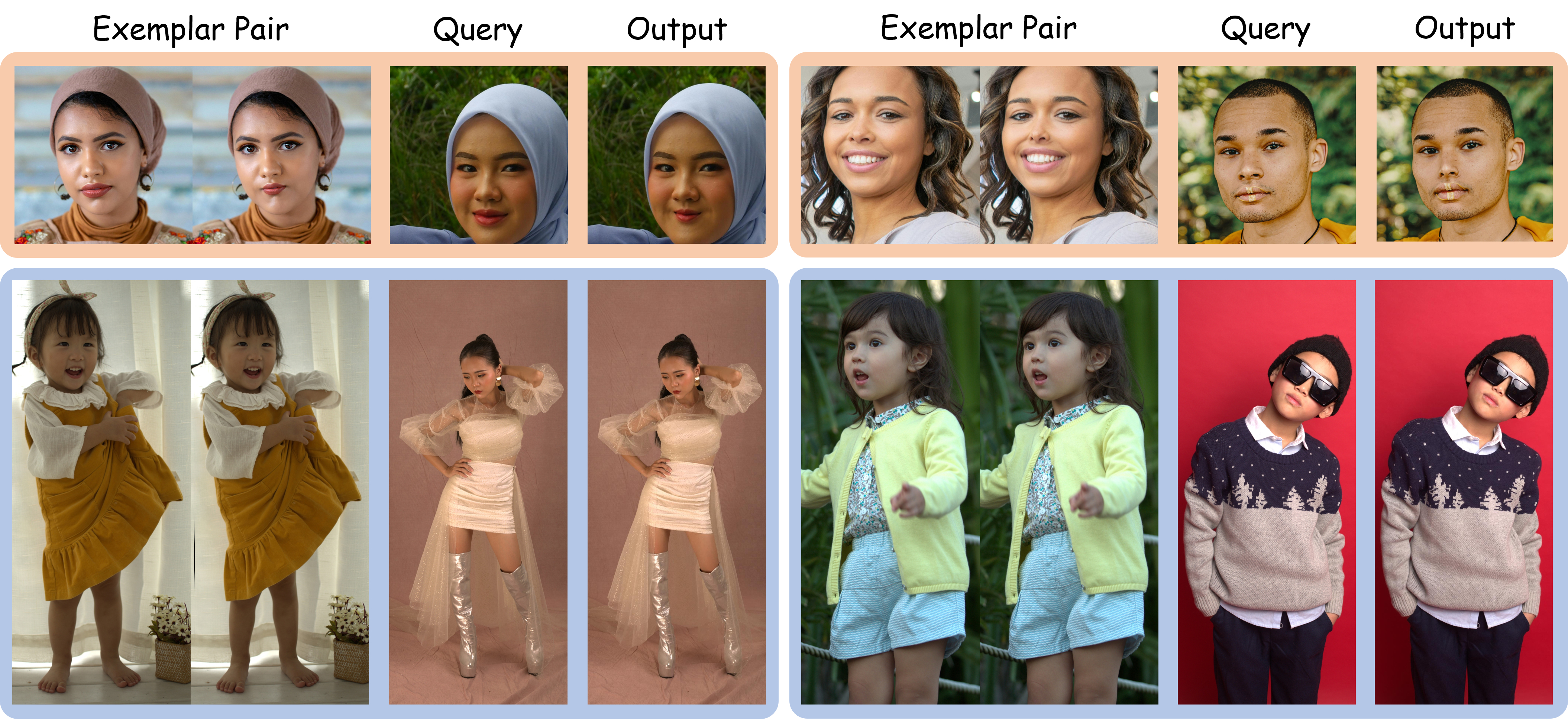}
    \caption{Results of MirrorPPR on various samples. MirrorPPR successfully extracts fine-grained retouching operations from the exemplar and transfers them to the query. The operation types for each sample are: (a) top-left: ``Shrink mouth''; (b) top-right: ``Enlarge eyes, Plump lips, Narrow nasal alae''; (c) bottom-left: ``Slim legs''; (d) bottom-right: ``Narrow nasal alae, Square shoulders''. }
    \label{fig:gallery}
\end{figure}

Text-guided image editing~\cite{LongCat-Image,wu2025omnigen2,wu2025qwenimagetechnicalreport,liu2025step1x-edit,flux-2-2025,brack2024leditslimitlessimageediting} has achieved remarkable success. However, it shows intrinsic limitations when applied to portrait photo retouching. Portrait retouching encompasses appearance-level (e.g., color grading, skin smoothing) and structural-level adjustments (e.g., refining jawlines, enlarging eyes, or modifying body proportions). While appearance retouching can be efficiently handled by existing parametric filters, structural retouching remains challenging and relies heavily on manual expertise. Natural language is inherently ambiguous for automating such tasks, as it struggles to quantitatively specify the exact spatial scale, direction, and magnitude of fine-grained structural manipulations. Consequently, text-guided models frequently suffer from misinterpretations, leading to insufficient adjustments or unnatural over-editing.

To bypass the expressive limitations of text, exemplar-based image editing~\cite{zhaoInstructBrushLearningAttentionbased2024a,yang2023imagebrush,xu2025textualize,srivastava2024reedit,nguyen2023visual,li2026viral,gong2025relationadapter,chen2025edit,lai2025unleashing,lu2025pairedit,wang2023context} has emerged as a promising alternative, allowing users to convey their intent intuitively via before-and-after image pairs. Yet, adapting this paradigm to structural portrait retouching introduces a unique challenge: unlike general image editing tasks characterized by substantial visual transformations, structural retouching operations are remarkably subtle. Consequently, existing exemplar-based models often lack the sensitivity to perceive and transfer such minute differences.

Furthermore, exemplar-based structural retouching suffers from a scarcity of suitable datasets. Existing datasets~\cite{li2018beautygan,shafaei2021autoretouch,cai2018learning,bharati2017demography,bharati2016detecting,rathgeb2020differential,rathgeb2020prnu} primarily focus on appearance-level adjustments or lack operation diversity. More critically, constructing cross-identity training pairs—where two distinct portraits undergo the exact same combination of retouching operations—is extremely difficult. Variations in image content, such as shot scales, partial occlusions, and head poses, mean that specific local edits are rarely universally applicable (e.g., applying leg thinning to a facial close-up, or modifying eye spacing on a profile face). Therefore, curating strictly paired, operation-aligned cross-identity datasets is unscalable and poses a critical data bottleneck.

To tackle these challenges, we introduce \textbf{MirrorPPR}, a framework that accurately captures and transfers subtle structural retouching operations. The framework consists of two modules. A Retouching Operation Extractor captures the subtle differences between the source and target exemplars, producing a retouching operation representation. This representation is then injected into a pre-trained Diffusion Transformer (DiT)~\cite{wu2025qwenimagetechnicalreport} backbone as a conditioning signal, guiding the DiT to transfer the structural retouching to a new query image. To optimize the framework, we employ a progressive two-stage training strategy: the extractor is first pre-trained, and subsequently linked to the DiT via a connector for joint LoRA~\cite{hu2022lora} fine-tuning.

To successfully optimize our framework and overcome the aforementioned cross-identity misalignment challenge, we propose an advanced data self-augmentation paradigm. Instead of using a different identity for the query, we construct the training quadruplet directly from the exemplar pair by applying a randomized set of spatial augmentations to the source and target exemplars synchronously. This self-augmented formulation ensures strict alignment of the retouching operations between the exemplar pair and the query-ground truth pair, effectively reducing optimization difficulty and accelerating model convergence.

Benefiting from the data self-augmentation strategy, we eliminate the need for strictly paired cross-identity examples, requiring only a diverse retouching dataset. To this end, we construct \textbf{MirrorPPR47M}, comprising over 47 million retouched pairs. Because authentic structural edits are very subtle to learn from scratch, we structure the dataset to support curriculum learning~\cite{bengio2009curriculum}: it includes a simulated subset with pronounced deformations for pre-training, and a professional subset for fine-tuning on highly realistic operations.

Extensive experiments validate that our approach significantly outperforms existing baselines, achieving state-of-the-art performance in both retouching quality and identity preservation.

In summary, our main contributions are threefold:
\begin{itemize}
\item We introduce the novel task of \textit{Exemplar-Based Portrait Photo Retouching} that focuses on structural reshaping, and propose \textbf{MirrorPPR}, an innovative framework that accurately captures and transfers subtle retouching operations.
\item We design an advanced data self-augmentation paradigm that effectively resolves the cross-identity misalignment challenge. 
\item We construct \textbf{MirrorPPR47M}, a large-scale dataset comprising over 47 million pairs for comprehensive structural portrait reshaping, establishing a solid foundation for mastering intricate real-world retouching operations.
\end{itemize}

\section{Related Work}

\noindent\textbf{Text-guided Image Editing.} Diffusion-based models~\cite{sohl2015deep,meng2021sdedit,song2019generative,ho2020denoising} have revolutionized text-guided image editing by leveraging natural language instructions~\cite{couairon2022diffeditdiffusionbasedsemanticimage, brack2024leditslimitlessimageediting, liu2025step1x-edit,wu2025omnigen2, LongCat-Image,sheynin2024emu}. Early works like P2P~\cite{hertz2022prompt} and InstructPix2Pix~\cite{brooks2023instructpix2pix} handle general edits, while MasaCtrl~\cite{cao2023masactrl} enables tuning-free non-rigid editing. Recent state-of-the-art models further support complex tasks like image composition and style transfer via multi-reference editing~\cite{flux-2-2025,wu2025qwenimagetechnicalreport,hurst2024gpt,team2023gemini,li2023instructany2pix,cai2026idglowdynamicidentitymodulation}. Despite their versatility, these methods remain inadequate for structural portrait retouching, as natural language lacks the precision to specify the exact spatial scale, direction, and magnitude required for subtle adjustments.

\noindent\textbf{Exemplar-Based Image Editing.} To address the semantic ambiguity of text, a growing body of research has shifted toward exemplar-based editing~\cite{yang2023imagebrush, gong2025relationadapter, lai2025unleashing,lu2025pairedit,wang2023context,srivastava2024reedit,liao2017visual,yang2023paint,gu2024analogist}. Early attempts cast this as an inversion problem, mapping visual examples to textual embeddings or discrete instructions~\cite{nguyen2023visual, zhaoInstructBrushLearningAttentionbased2024a, xu2025textualize}. Recent works advance toward In-Context Learning (ICL), with methods like EditTransfer~\cite{chen2025edit} and VIRAL~\cite{li2026viral} addressing non-rigid transformations and heterogeneous tasks. However, as these frameworks largely target general-purpose editing, extending them to extract and transfer the fine-grained edits required for structural portrait retouching remains an underexplored challenge.

\noindent\textbf{Portrait Photo Retouching.} Traditional portrait retouching pipelines~\cite{cai2018learning, hu2018exposure, kim1997contrast,kosugi2020unpaired,mantiuk2008display,mukherjee2008enhancement} primarily focus on global, appearance-level enhancements like color grading and skin smoothing. While generative models such as StyleGAN~\cite{karras2019style,tewari2020stylerig,medin2022most} enable structural facial editing, they still struggle with fine-grained, localized adjustments. A major bottleneck for AI-driven structural retouching is the lack of suitable training data. Existing datasets are either dominated by appearance adjustments~\cite{bychkovsky2011learning,shafaei2021autoretouch,li2018beautygan} or lack operation diversity~\cite{bharati2016detecting, rathgeb2020prnu,rathgeb2020differential}.
To resolve these issues, we carefully construct a large-scale structural portrait retouching dataset.

\section{Method}

\subsection{Task Formulation and Method Overview}
\label{sec:formulation_overview}

Given an exemplar pair $(X_s, X_t)$ representing the source and its retouched counterpart, alongside a new query image $X_q$, the objective of exemplar-based portrait photo retouching is to generate a target image $\hat{Y}_q$ by applying the identical retouching operations demonstrated in the exemplar pair. The ground-truth retouched image corresponding to $X_q$ is denoted as $Y_q$.

\begin{figure}[t]
    \centering
    \includegraphics[width=1\linewidth]{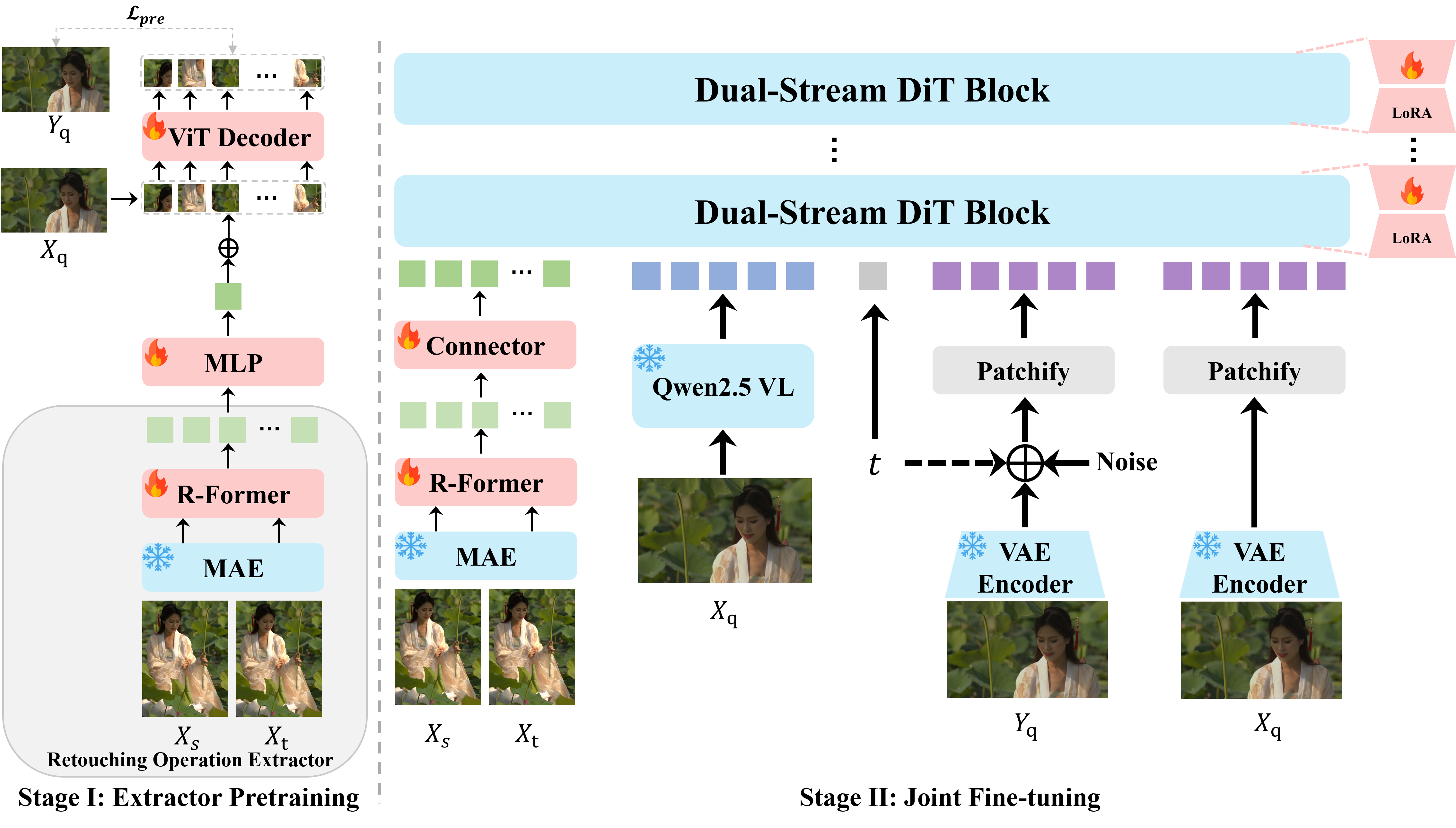}
    \caption{Overall architecture of the proposed MirrorPPR framework. Our framework is trained in a progressive two-stage pipeline. Left: A Retouching Operation Extractor, comprising a frozen MAE and a trainable R-Former, extracts subtle retouching operations from the exemplar pair $(X_s, X_t)$. It is pre-trained via an auxiliary reconstruction task using a temporary MLP and ViT decoder. Right: The pre-trained R-Former is integrated with a frozen dual-stream DiT. The extracted operation features are processed by a trainable connector and injected into the DiT blocks. The R-Former, connector, and newly added LoRA modules are jointly fine-tuned to transfer the operations. The snowflake and fire icons denote frozen and trainable modules, respectively.}
    \label{fig:architecture}
\end{figure}

As shown in Figure~\ref{fig:architecture}, our proposed MirrorPPR framework consists of two core components: (1) a \textbf{Retouching Operation Extractor} that captures subtle retouching operations from the exemplar pair, and (2) a pre-trained \textbf{Diffusion Transformer (DiT)} backbone that transfers these operations to the query image via a dedicated connector and LoRA modules.

\subsection{Retouching Operation Extractor}
\label{sec:operation_extraction}

Portrait retouching involves extremely subtle modifications. Inspired by Moto~\cite{chen2025moto}, we introduce the Retouching Operation Extractor to accurately capture subtle retouching operations. It consists of a frozen Masked Autoencoder (MAE)~\cite{he2022masked} and a Transformer-based network, termed \textbf{R-Former}, which is specifically designed to extract retouching operations from the MAE features. MAE is particularly suitable for this extractor, as it preserves local spatial structures and provides representations favorable to geometry, localization, and fine-grained differences~\cite{xie2023revealing}, thereby facilitating the capture of subtle retouching operations.

\noindent\textbf{Retouching Operation Extraction.} The patch-level features extracted by the frozen MAE from the exemplar pair are passed into the R-Former. Within the network, a set of internal learnable query tokens is concatenated with the incoming image features along the sequence dimension. As they pass through the self-attention layers, the query tokens interact with the image features, enabling them to extract the subtle retouching operations from the image features. We retain only the output representations corresponding to these queries, denoted as $\mathbf{H}_{edit}$, which serve as the raw retouching operation representation.

\noindent\textbf{Extractor Pre-training.} To ensure that $\mathbf{H}_{edit}$ accurately encapsulates the precise retouching intent, we introduce an auxiliary reconstruction proxy task to pre-train the extractor. Specifically, a Multi-Layer Perceptron (MLP) projects $\mathbf{H}_{edit}$ into a compact edit embedding $\mathbf{e}_{edit}$. For a given query image $X_q$, its patch embeddings are extracted, and $\mathbf{e}_{edit}$ is added to each patch token. A lightweight ViT decoder then processes these fused tokens to reconstruct the retouched result $\hat{Y}_q$. This pre-training stage is optimized using a combination of Mean Squared Error (MSE) reconstruction loss and LPIPS perceptual loss~\cite{zhang2018unreasonable}:
\begin{equation}
    \mathcal{L}_{pre} = \|\hat{Y}_q - Y_q\|_2^2 + \lambda \mathcal{L}_{lpips}(\hat{Y}_q, Y_q),
\end{equation}
where $\lambda$ is a balancing scalar. After pre-training, the extractor learns robust operation representations, and the auxiliary MLP and ViT decoder are discarded.

\subsection{Operation Transfer via Diffusion Transformer}
\label{sec:operation_transfer}

After pre-training the extractor, we integrate it with a pre-trained image editing diffusion model to apply the extracted retouching operations to the query image.

\noindent\textbf{Backbone.} We adopt Qwen-Image-Edit-2511~\cite{wu2025qwenimagetechnicalreport} as our backbone, which features a dual-stream DiT.
One stream leverages the Multimodal Large Language Model (MLLM) Qwen2.5-VL~\cite{bai2025qwen25vltechnicalreport} to extract high-level visual representations, while the other stream performs diffusion denoising on the target latents conditioned on image VAE latents.
Since our exemplar-based task is fundamentally driven by visual demonstrations rather than textual descriptions, we omit text instructions entirely. We retain the frozen Qwen2.5-VL to extract the visual feature embeddings $\mathbf{c}_{img}$ from $X_q$, and preserve the frozen VAE encoder $\mathcal{E}$ to obtain the image latent representation $\mathbf{z}_{cond} = \mathcal{E}(X_q)$.

\noindent\textbf{Operation Injection.} To inject the retouching operations into the diffusion process, the extracted operation representations $\mathbf{H}_{edit}$ are processed by a trainable connector. The connector maps $\mathbf{H}_{edit}$ into the instruction conditioning space of the DiT backbone, producing the final retouching operation condition $\mathbf{c}_{edit}$.

\noindent\textbf{Joint Fine-tuning.} Let $\mathbf{z}_0 = \mathcal{E}(Y_q)$ be the encoded ground-truth target latent, and $\mathbf{z}_1 \sim \mathcal{N}(\mathbf{0}, \mathbf{I})$ be a standard Gaussian noise vector. For a given timestep $t \in [0, 1]$, the intermediate noisy latent $\mathbf{z}_t$ is constructed via linear interpolation, and the target velocity $\mathbf{v}_t$ is defined as the trajectory from noise to data:
\begin{equation}
    \mathbf{z}_t = t \mathbf{z}_0 + (1-t) \mathbf{z}_1, \quad \quad \mathbf{v}_t = \mathbf{z}_0 - \mathbf{z}_1.
\end{equation}

The flow matching loss is formulated as the MSE between the predicted velocity $\mathbf{v}_\theta$ and the ground-truth velocity $\mathbf{v}_t$:
\begin{equation}
    \mathcal{L}_{flow} = \mathbb{E}_{\mathbf{z}_1, \mathbf{z}_0, t} \left[ \left\| \mathbf{v}_\theta([\mathbf{z}_t, \mathbf{z}_{cond}], t, \mathbf{c}_{img}, \mathbf{c}_{edit}) - \mathbf{v}_t \right\|_2^2 \right].
\end{equation}

To efficiently adapt the powerful prior of the pre-trained DiT to our task, we freeze its original parameters and add trainable LoRA~\cite{hu2022lora} modules into its attention blocks. The pre-trained R-Former, the connector, and the LoRA modules are end-to-end jointly optimized via $\mathcal{L}_{flow}$, enabling the network to accurately transfer the subtle retouching operations.

\subsection{Data Self-Augmentation Paradigm}
\label{sec:self_augmentation}

\begin{figure}[t]
    \centering
    \includegraphics[width=1\linewidth]{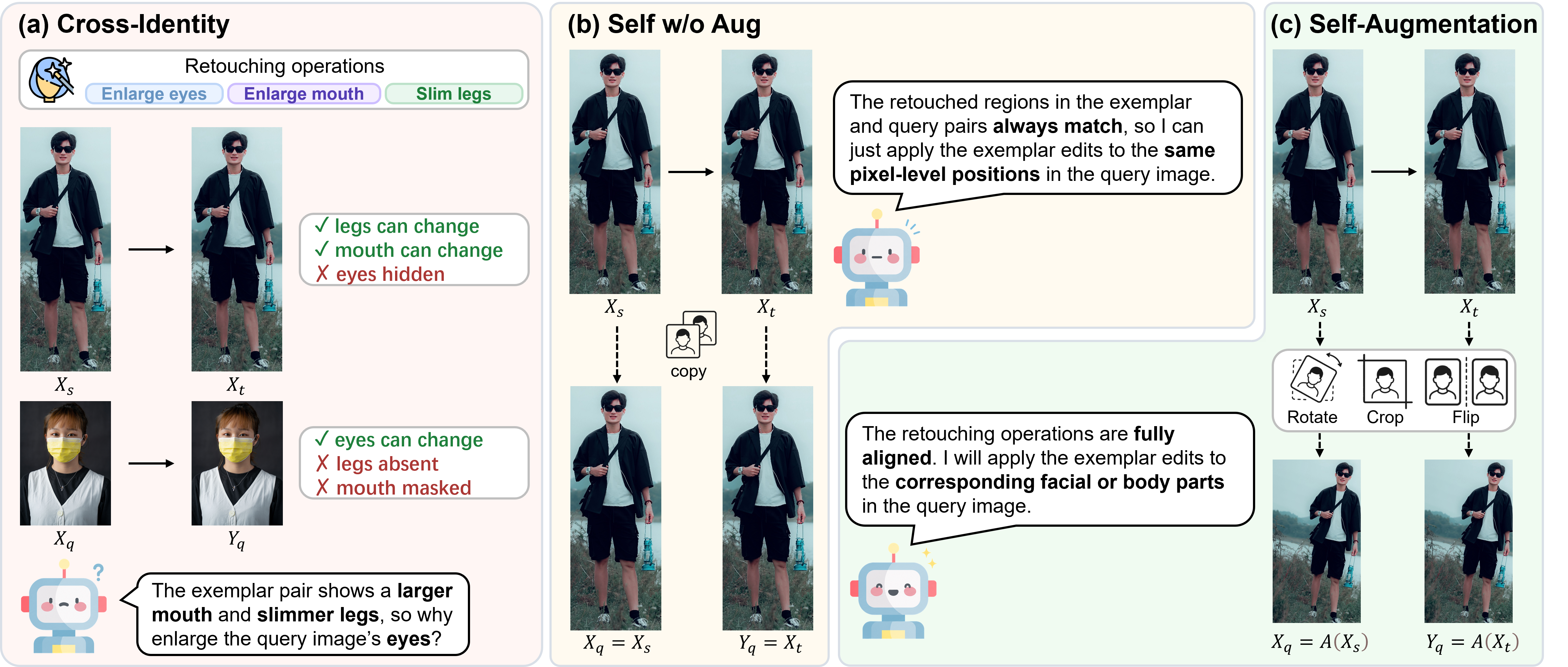}
    \caption{
Illustration of the proposed data self-augmentation paradigm.
The Cross-Identity setting suffers from operation misalignment between the exemplar pair and the query pair, while Self w/o Aug causes pixel-level shortcut learning.
Our Self-Augmentation applies the same spatial augmentation $A$ to both $X_s$ and $X_t$, which preserves operation alignment while breaking absolute coordinate correspondence.
}
    \label{fig:self_aug}
\end{figure}

As discussed in Section~\ref{sec:intro}, constructing valid training quadruplets for exemplar-based portrait retouching is non-trivial.
A straightforward strategy is to build \textit{Cross-Identity} quadruplets, where the exemplar pair $(X_s, X_t)$ and query pair $(X_q, Y_q)$ come from different identities and are expected to share the same intended retouching operations.
However, as shown in Figure~\ref{fig:self_aug} (a), due to variations in shot scale, pose, occlusion, and portrait composition, the operation between the two pairs may not strictly correspond.
Such misalignment makes the supervision ambiguous and hinders the model from learning a consistent operation-transfer mapping from the exemplar pair to the query pair.

A natural workaround is to construct the training quadruplet directly from a single exemplar pair.
We denote the naive version as \textit{Self w/o Aug}, where the query image is set to the source exemplar and the ground truth is set to the target exemplar, i.e., $X_q=X_s$ and $Y_q=X_t$.
Although this construction avoids cross-identity operation misalignment, it introduces severe \textit{shortcut learning}.
Since the exemplar pair and the query pair are perfectly aligned at the pixel level, the model can trivially copy coordinate-wise differences rather than infer the underlying retouching semantics.
Consequently, it struggles to generalize to query images with different identities, spatial layouts, or scales.

To break this spatial coupling while retaining exact operation consistency, we introduce \textit{Self-Augmentation}.
Specifically, we synthesize the query pair by applying a shared random spatial augmentation $A$, such as scaling, cropping, rotation, and horizontal flipping, to both the source and target images.
The query pair is therefore constructed as $X_q=A(X_s)$ and $Y_q=A(X_t)$.
Because the same augmentation is applied to both images, the retouching operations in the synthesized query pair remain consistent with those in the exemplar pair.
Meanwhile, the transformed query pair no longer shares the same absolute pixel coordinates with the original exemplar pair $(X_s, X_t)$, preventing shortcut learning.
By preserving operation alignment while breaking absolute pixel-level correspondence, this construction encourages the model to transfer the retouching operations in the exemplar pair according to the query image's own spatial layout, thereby overcoming the cross-identity data bottleneck and enabling robust generalization of the learned operations.

\section{MirrorPPR47M Dataset}
\label{sec:dataset}

To overcome the scarcity of paired geometric retouching data, we construct \textbf{MirrorPPR47M}, a large-scale portrait dataset encompassing comprehensive structural reshaping. These operations extensively cover facial features, face contours, and body proportions across diverse shot scales. Since real-world retouching is very subtle, training a network from scratch is highly challenging. To mitigate this, MirrorPPR47M is designed to support an easy-to-hard \textit{curriculum learning strategy}. It comprises a \textbf{Simulated Retouching Subset} featuring pronounced deformations and a \textbf{Professional Retouching Subset} providing authentic, complex operations. We construct the dataset through the following pipeline.

\subsection{Data Curation and Filtering}
We source raw high-resolution images and apply strict filtering criteria based on head pose, facial occlusion, and portrait area ratio. Detailed criteria are provided in Appendix~\ref{app:Data Filtering Criteria}. For the \textit{Simulated Retouching Subset}, we retain $30,171$ high-quality images from the FFHQ dataset~\cite{karras2019style}. For the \textit{Professional Retouching Subset}, we select $3,789$ 4K-8K portraits from the PPR10K dataset~\cite{liang2021ppr10k}. 

\subsection{Retouching Execution}
\noindent\textbf{Simulated Retouching Generation.} To synthesize pronounced geometric deformations, we propose the Landmark-Guided Local Warping (LLW) algorithm.
Detailed formulations and implementations of this algorithm are provided in Appendix~\ref{app:LLW}. We simulate 8 types of facial operations, each with two opposite directions. By randomly combining 1 to 8 operations per identity, we synthesize $808,439$ retouched pairs.

\noindent\textbf{Professional Retouching Generation.} To approximate real-world authenticity, we process the source images from PPR10K using a commercial retouching API. We select 27 professional operations, comprising 18 for facial features, 4 for face shapes, and 5 for body proportions. Applying random combinations of 1 to 7 operations per image yields $46,642,845$ finely retouched pairs.

Appendix~\ref{app:supp_operation_taxonomy_statistics} provides more detailed operation taxonomy and statistics.

\subsection{Data Processing and Self-Augmentation}

Following the Self-Augmentation paradigm in Section~\ref{sec:self_augmentation}, we implement a concrete data processing pipeline to generate spatially decoupled yet operation-aligned training quadruplets $(X_s, X_t, X_q, Y_q)$.

We first crop and resize raw images to a total area of approximately 4 MP. This cropping process utilizes facial bounding boxes and human body masks to strictly ensure the completeness of the portrait subject. Detailed cropping procedures are provided in Appendix~\ref{app:data process}. Subsequently, we synchronously apply randomized spatial augmentations to the source and target images. These augmentations include mirroring, rotation, and dynamic cropping with varied spatial offsets. Consequently, the retouching operations in the query pair $(X_q, Y_q)$ appear at entirely different spatial coordinates and scales compared to the exemplar pair $(X_s, X_t)$. This strategy effectively prevents shortcut learning and expands each original pair into approximately 13.3 spatially decoupled variations.

\subsection{Progressive Curriculum Learning}
\label{sec:4.4}
We use the two subsets to design a progressive training strategy. First, the Retouching Operation Extractor is pre-trained on the Simulated Retouching Subset to capture fundamental structural variations, then further trained on the Professional Retouching Subset to adapt to intricate, real-world retouching subtleties. In the final fine-tuning stage, all trainable modules are exclusively optimized on the professional data. This tailored data recipe ensures that the network can stably capture highly complex structural retouching operations.
\section{Experiments}
\label{sec:experiments}

\subsection{Experimental Setting}
\label{sec:exp_setting}

\subsubsection{Benchmarks.}
To rigorously evaluate cross-identity operation transfer capabilities, we construct two benchmarks where the exemplar pair and query image feature entirely different identities—a crucial distinction from our self-augmented training phase. \textbf{SimFace-100} comprises 100 retouching combinations randomly paired from 12 distinct $1024 \times 1024$ facial images using 8 operations supported by LLW, testing the preliminary perception of prominent simulated retouching. \textbf{ProPortrait-500} involves 500 combinations of 27 professional operations applied to 40 high-quality portraits using the same commercial API as the Professional Retouching Subset, with images standardized to an area equivalent to 1.5K \(\times\) 1.5K pixels. This benchmark poses a severe challenge for capturing highly subtle and fine-grained professional edits.

\subsubsection{Evaluation Metrics.}
We use PSNR, SSIM~\cite{wang2004image}, and LPIPS~\cite{zhang2018unreasonable} to evaluate pixel-level fidelity, structural reconstruction accuracy, and perceptual similarity, respectively. Following MoFRR~\cite{liu2025mofrr}, we introduce Face Similarity to assess identity preservation when retouching. This metric leverages the face recognition model ArcFace~\cite{deng2019arcface} to extract feature embeddings from the output and the ground truth, and computes their cosine similarity for assessment.

\subsubsection{Implementation Details.}
Corresponding to our benchmarks, we train two model variants. \textbf{MirrorPPR-Face} is trained in two stages on $1024 \times 1024$ simulated facial crops extracted using FFHQ facial bounding boxes. \textbf{MirrorPPR-Pro} undergoes the progressive curriculum learning proposed in Section~\ref{sec:4.4}. Specifically, the extractor is sequentially pre-trained on the simulated and professional subsets, followed by joint fine-tuning of the entire framework on professional data. Detailed network configurations and training settings are provided in Appendices~\ref{app:arch_details} and~\ref{app:training_details}.

\subsection{Comparison with Other Methods}

To comprehensively evaluate the effectiveness of our proposed MirrorPPR, we compare it against powerful baselines across three distinct categories. For multi-reference image editing, we choose Qwen-Image-Edit-2511~\cite{wu2025qwenimagetechnicalreport}, FLUX.2-dev~\cite{flux-2-2025}, Nano Banana 2~\cite{team2023gemini}, and Seedream 4.5~\cite{seedream2025seedream}. For exemplar-based image editing, we select three recent DiT-based methods: Qwen-Image-Edit-2511-ICEdit-LoRA~\cite{qwen_image_edit_2024}, RelationAdapter~\cite{gong2025relationadapter}, and EditTransfer~\cite{chen2025edit}. Additionally, we evaluate the text-guided editing capabilities of Qwen-Image-Edit-2511, FLUX.2-dev, Nano Banana 2, and Seedream 4.5. For multi-reference and exemplar-based methods, we uniformly use the prompt: "Infer the portrait retouching operations applied from Image 1 (original) to Image 2 (retouched), then apply the same retouching operations to Image 3." For text-guided methods, we provide clear text instructions. Additional evaluation details are provided in Appendix~\ref{app:eval_details}.

\subsubsection{Quantitative Analysis.}

\begin{table*}[t]
\centering
\caption{Quantitative comparison results with baselines on SimFace-100.}
\label{tab:comparison_results_second}
\resizebox{\textwidth}{!}{ 
\begin{tabular}{llcccc}
\toprule
\textbf{Category} & \textbf{Model} & \textbf{PSNR} $\uparrow$ & \textbf{SSIM} $\uparrow$ & \textbf{LPIPS} $\downarrow$ & \textbf{Face Similarity} $\uparrow$ \\ 
\midrule

\multirow{4}{*}{\begin{tabular}[c]{@{}l@{}}Multi-reference \\ Image Editing\end{tabular}} 
& Qwen-Image-Edit-2511 & 9.06 & 0.468 & 0.745 & 0.207 \\
& FLUX.2-dev & 9.28 & 0.481 & 0.698 & 0.110 \\
& Nano Banana 2 & 16.72 & 0.784 & 0.329 & 0.556 \\
& Seedream 4.5 & 13.01 & 0.709 & 0.501 & 0.351 \\

\midrule
\multirow{3}{*}{Exemplar-based} 
& Qwen-Image-Edit-2511-ICEdit-LoRA & 9.21 & 0.533 & 0.640 & 0.300 \\
& RelationAdapter & 16.57 & 0.698 & 0.543 & 0.204 \\
& EditTransfer & 15.68 & 0.691 & 0.492 & 0.464 \\

\midrule
\multirow{4}{*}{Text-guided} 
& Qwen-Image-Edit-2511 & 25.80 & 0.862 & 0.260 & 0.463 \\
& FLUX.2-dev & 22.44 & 0.804 & 0.301 & 0.531 \\
& Nano Banana 2 & 24.25 & 0.860 & 0.239 & 0.601 \\
& Seedream 4.5 & 18.01 & 0.788 & 0.368 & 0.600 \\

\midrule
\multirow{1}{*}{\textbf{Ours}} 
& \textbf{MirrorPPR-Face} & \textbf{32.25} & \textbf{0.909} & \textbf{0.186} & \textbf{0.937} \\

\bottomrule
\end{tabular}
} 
\end{table*}

\begin{table*}[t]
\centering
\caption{Quantitative comparison results with baselines on ProPortrait-500.}
\label{tab:comparison_results_ppr}
\resizebox{\textwidth}{!}{ 
\begin{tabular}{llcccc}
\toprule
\textbf{Category} & \textbf{Model} & \textbf{PSNR} $\uparrow$ & \textbf{SSIM} $\uparrow$ & \textbf{LPIPS} $\downarrow$ & \textbf{Face Similarity} $\uparrow$ \\ 
\midrule

\multirow{4}{*}{\begin{tabular}[c]{@{}l@{}}Multi-reference \\ Image Editing\end{tabular}} 
& Qwen-Image-Edit-2511 & 10.23 & 0.538 & 0.645 & 0.413 \\
& FLUX.2-dev & 9.36 & 0.466 & 0.728 & 0.220 \\
& Nano Banana 2 & 17.72 & 0.835 & 0.250 & 0.811 \\
& Seedream 4.5 & 12.12 & 0.689 & 0.436 & 0.705 \\

\midrule
\multirow{3}{*}{Exemplar-based} 
& Qwen-Image-Edit-2511-ICEdit-LoRA & 12.06 & 0.631 & 0.564 & 0.606 \\
& RelationAdapter & 15.74 & 0.709 & 0.586 & 0.283 \\
& EditTransfer & 18.32 & 0.748 & 0.481 & 0.457 \\

\midrule
\multirow{4}{*}{Text-guided} 
& Qwen-Image-Edit-2511 & 20.85 & 0.732 & 0.387 & 0.501 \\
& FLUX.2-dev & 19.94 & 0.748 & 0.345 & 0.616 \\
& Nano Banana 2 & 27.45 & 0.904 & \textbf{0.183} & 0.667 \\
& Seedream 4.5 & 16.43 & 0.770 & 0.378 & 0.782 \\

\midrule
\multirow{1}{*}{\textbf{Ours}} 
& \textbf{MirrorPPR-Pro} & \textbf{32.65} & \textbf{0.927} & 0.200 & \textbf{0.960} \\

\bottomrule
\end{tabular}
} 
\end{table*}

Quantitative results on the SimFace-100 and ProPortrait-500 benchmarks are summarized in Table~\ref{tab:comparison_results_second} and Table~\ref{tab:comparison_results_ppr}, respectively. From the results, we draw the following conclusions:

\textbf{Multi-reference and exemplar-based methods completely fail on this task.} 
As shown in the tables, existing multi-reference editing models (with the exception of Nano Banana 2) and baseline exemplar-based methods exhibit extremely poor performance. All evaluation metrics fall far below the normal ranges expected for portrait retouching tasks.

\textbf{Text-guided methods perform better in reconstruction but struggle with identity preservation.} 
Compared to the first two categories, text-guided models demonstrate significantly better pixel-level reconstruction and structural fidelity, achieving much higher PSNR and SSIM scores. However, a prominent weakness across these models is their noticeably low Face Similarity scores. For instance, although the text-guided Nano Banana 2 achieves a highly competitive LPIPS of 0.183 on the ProPortrait-500 benchmark, its Face Similarity drops to merely 0.667. This sharp contrast indicates that text-guided models severely compromise the biometric characteristics of the portrait subject.

\textbf{MirrorPPR demonstrates comprehensive superiority.} 
Both MirrorPPR-Face and MirrorPPR-Pro consistently yield excellent scores across the evaluation metrics on both benchmarks. On the highly challenging ProPortrait-500 dataset, MirrorPPR-Pro achieves a PSNR of 32.65, an SSIM of 0.927, and an unprecedented Face Similarity of 0.960. These superior metrics robustly demonstrate the effectiveness of our method for exemplar-based portrait retouching.

\subsubsection{Qualitative Comparison.}

\begin{figure}[t]
\centering
\includegraphics[width=\linewidth]{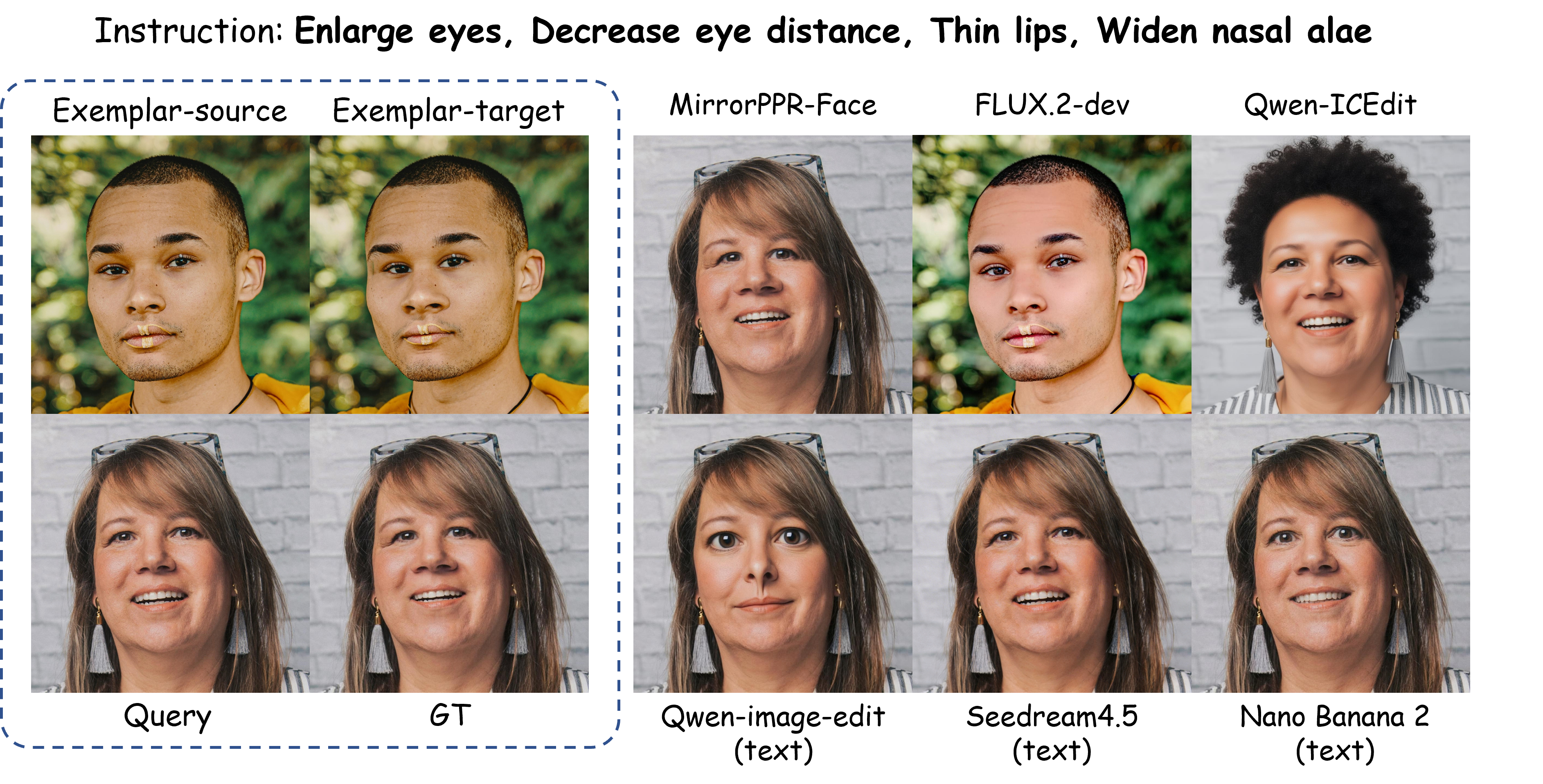}
\caption{Qualitative comparison on SimFace-100. MirrorPPR-Face accurately captures and faithfully transfers all operations, while other models suffer from various issues.}
\label{fig:qualitative}
\end{figure}

\begin{figure}[t]
    \centering
    \includegraphics[width=\linewidth]{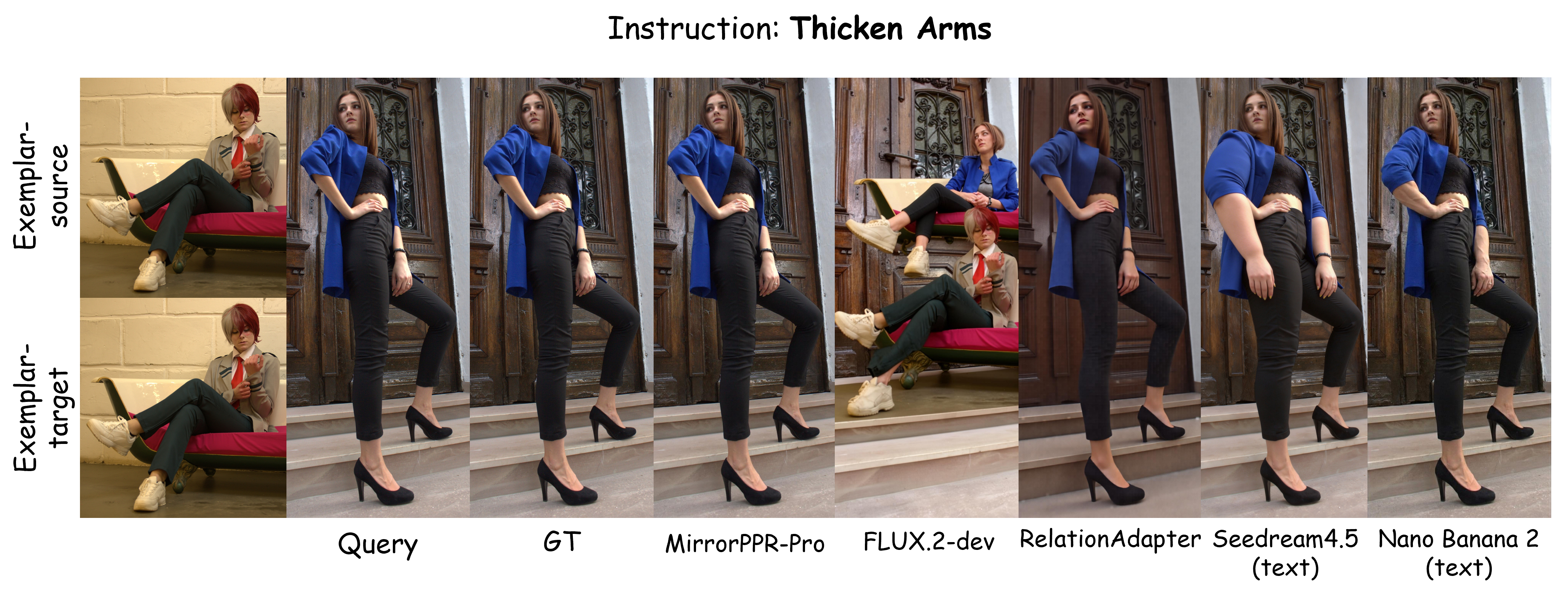}
    \caption{Qualitative comparison on ProPortrait-500. MirrorPPR-Pro accurately captures and transfers all operations, while other models suffer from various issues.}
    \label{fig:qulitative2}
\end{figure}

We manually checked the test outputs to explain the metrics and summarize the causes of baseline failures.
Specifically, multi-reference and exemplar-based models typically misinterpret operation transfer as image blending or face swapping. Meanwhile, text-guided models lack precise geometric control due to language ambiguity, frequently producing anatomically exaggerated, "over-edited" features that corrupt the original identity. In contrast, our MirrorPPR achieves precise structural reshaping while perfectly preserving the unedited details. Representative examples illustrating these observations are shown in Figure~\ref{fig:qualitative} and Figure~\ref{fig:qulitative2}.

\subsubsection{User Study.}
To complement automatic metrics and visual comparisons, we further conduct a user study in which participants select the anonymized candidate that best transfers the demonstrated retouching operation while preserving identity and avoiding changes to unrelated regions. MirrorPPR-Pro receives 79.0\% overall preference, showing that its advantage is also clearly perceived by human users. More details about the user study are provided in Appendix~\ref{app:user_study}.

\subsection{Latent Space Analysis of the Extracted Operations}
\label{sec:analysis}

To gain a deeper understanding of the operation representations learned by our framework, we conduct further analyses on the edit embeddings $\mathbf{e}_{edit}$. These embeddings are obtained from the Retouching Operation Extractor and subsequently mapped through the auxiliary MLP. We evaluate these embeddings on the SimFace-100 benchmark, revealing several compelling properties of the learned latent space, including robust operation transfer consistency, distinct operation clustering, and vector additivity.

\noindent\textbf{Operation Transfer Consistency.}
A key requirement of our framework is to ensure that the exact retouching intent demonstrated in the exemplar pair is faithfully transferred to the target query image. To quantitatively verify this, we feed the exemplar pair $(X_s, X_t)$ into the pre-trained extractor and the MLP to obtain the edit embedding $\mathbf{e}_{edit}^{exemplar}$. Concurrently, we extract the edit embedding $\mathbf{e}_{edit}^{query}$ from the query image and our generated retouching result $(X_q, \hat{Y}_q)$. 

On SimFace-100, the average cosine similarity between $\mathbf{e}_{edit}^{exemplar}$ and $\mathbf{e}_{edit}^{query}$ is remarkably high at 0.950. This exceptionally high similarity rigorously ensures that the retouching operations in the output perfectly align with the exemplar.

\begin{figure}[t]
    \centering
    \begin{minipage}[t]{0.5\textwidth}
        \centering
        \includegraphics[width=1\linewidth]{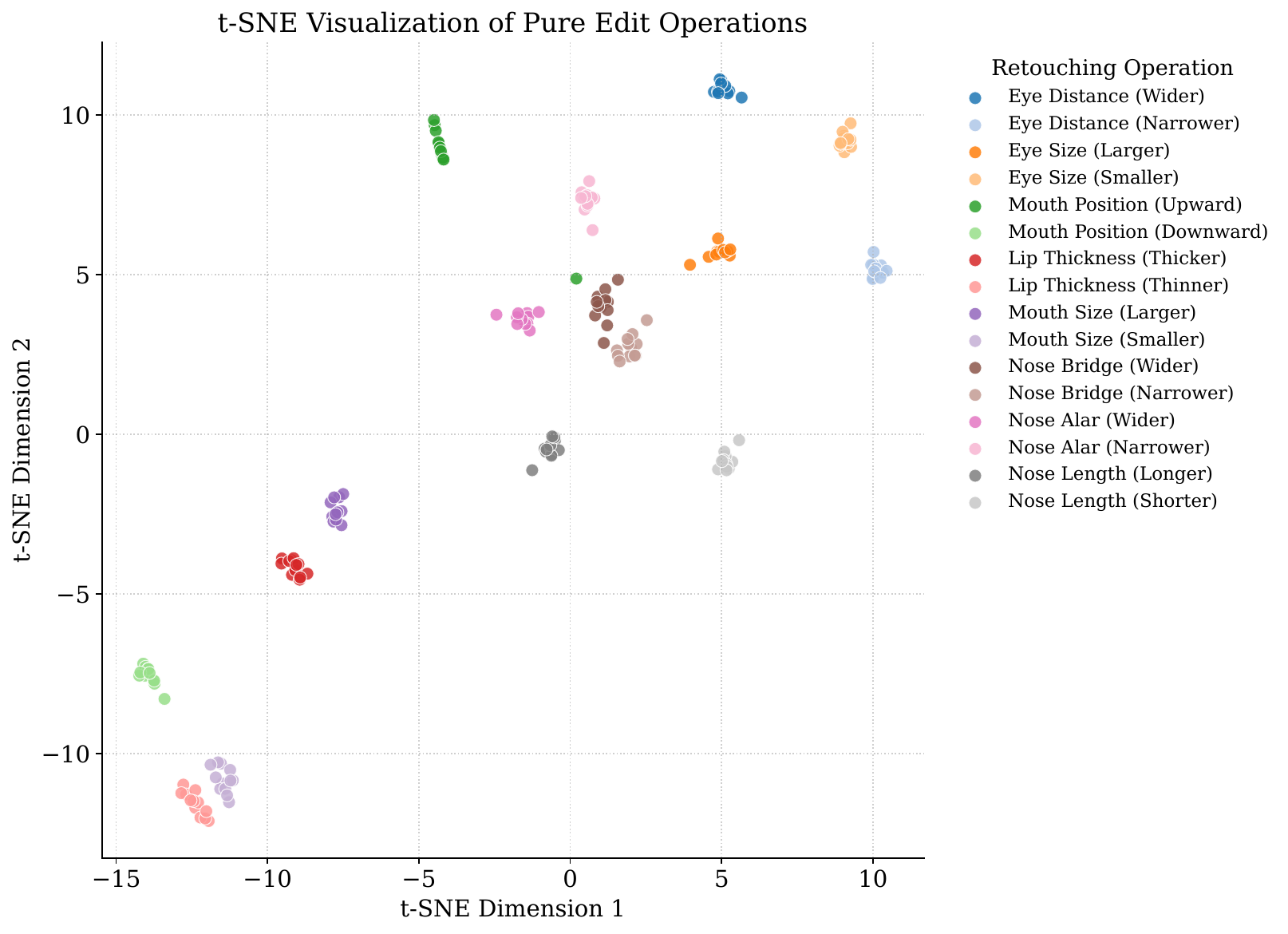}
        \caption{\textbf{t-SNE visualization of the retouching embeddings.} The clear clustering confirms our extractor's ability to learn distinct, identity-agnostic representations for different retouching operations.}
        \label{fig:tsne}
    \end{minipage}
    \hfill
    \begin{minipage}[t]{0.48\textwidth}
        \centering
        \includegraphics[width=\linewidth]{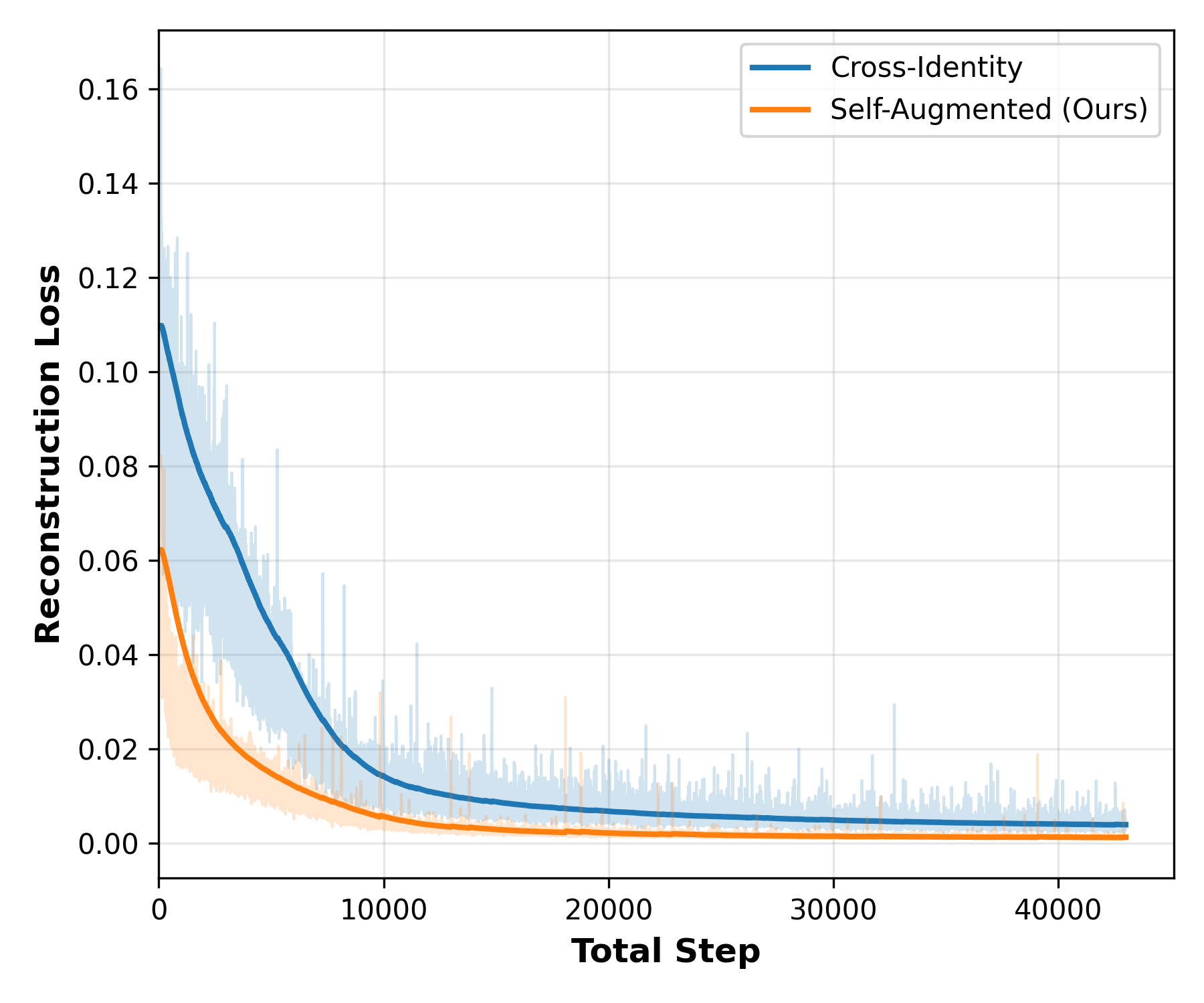} 
        \caption{Training loss curves of the Cross-Identity and Self-Augmentation settings.}
        \label{fig:ablation_curve}
    \end{minipage}
\end{figure}

\noindent\textbf{Clustering of Retouching Operations.}
We further investigate whether our pre-trained extractor learns distinct representations for different retouching operations. We construct a complete single-operation subset comprising 192 image pairs (12 distinct face images, where 16 different simulated single-operation edits are applied to each face) and extract their corresponding edit embeddings $\mathbf{e}_{edit}$. 

Raw edit embeddings from the pre-trained model inevitably retain instance-specific image characteristics. To isolate the pure retouching directions, we apply \textit{instance-level mean centering}: for each image, we subtract the mean vector of its 16 operations from the individual embeddings. As shown in Figure~\ref{fig:tsne}, these centered directional vectors exhibit clear operation-based clustering. 

\noindent\textbf{Vector Additivity.}
Based on the well-clustered embeddings, we further examine their additivity by testing whether they support direct vector addition for multiple operations. Using the identical 192-sample single-operation dataset, we attempt to composite multiple operations in the latent space. 

For a specific image requiring a composite edit (e.g., Operation A + Operation B), we first isolate the pure editing directions, denoted as $\Delta_{A}$ and $\Delta_{B}$, by applying the mean centering. The composite edit embedding is then computed as $\mathbf{e}_{composite} = \mathbf{e}_{mean} + \Delta_{A} + \Delta_{B}$, where $\mathbf{e}_{mean}$ is the mean base vector of that instance. This embedding is then directly fed into the pre-trained ViT decoder to generate the retouched result. 

\begin{table}[t]
\centering
\caption{Quantitative evaluation of vector additivity on SimFace-100. Simply adding the individual operation vectors directly in the latent space surpasses all baselines.}
\label{tab:additivity}
\resizebox{0.9\textwidth}{!}{
\begin{tabular}{lcccc}
\toprule
\textbf{Method} & \textbf{PSNR $\uparrow$} & \textbf{SSIM $\uparrow$} & \textbf{LPIPS $\downarrow$} & \textbf{Face Similarity $\uparrow$} \\
\midrule
MirrorPPR-Face & 32.25 & 0.909 & 0.186 & 0.937 \\
Latent Vector Additivity & 30.85 & 0.892 & 0.196 & 0.869 \\
\bottomrule
\end{tabular}
}
\end{table}

We evaluate this vector addition approach on SimFace-100. As presented in Table~\ref{tab:additivity}, it surpasses all compared baselines in Table~\ref{tab:comparison_results_second}. This compelling finding suggests that our learned latent space possesses a robust additive property, providing a reliable foundation for representing complex combined operations.

\subsection{Ablation Study}

We conduct two ablation studies to validate the proposed Data Self-Augmentation paradigm.
The first study tests whether training with Self-Augmentation causes a training-inference mismatch and weakens cross-identity operation transfer.
The second study investigates whether Self-Augmentation brings clearer benefits when cross-identity training data is difficult to keep operation-aligned.

\textbf{Self-Augmentation does not weaken cross-identity transfer.}
To verify this, we conduct an ablation on the simulated retouching subset and evaluate the models on SimFace-100.
We evaluate the three data construction strategies defined in Section~\ref{sec:self_augmentation} and illustrated in Figure~\ref{fig:self_aug}: Cross-Identity, Self w/o Aug, and Self-Augmentation.
In this ablation, the Cross-Identity setting is a relatively ideal scenario with good operation alignment, since the data are face-centric and the retouching operations are controlled by LLW.

\begin{table}[t]
\centering
\caption{Ablation results on SimFace-100 for evaluating whether training with Self-Augmentation weakens cross-identity transfer.}
\label{tab:sim_self_aug_ablation}
\begin{tabular}{lcccc}
\toprule
Setting & PSNR $\uparrow$ & SSIM $\uparrow$ & LPIPS $\downarrow$ & Face Similarity $\uparrow$ \\
\midrule
Self w/o Aug & 28.63 & 0.906 & 0.213 & 0.680 \\
Cross-Identity & 32.08 & \textbf{0.909} & \textbf{0.186} & \textbf{0.937} \\
Self-Augmentation (Ours) & \textbf{32.25} & \textbf{0.909} & \textbf{0.186} & \textbf{0.937} \\
\bottomrule
\end{tabular}
\end{table}

As shown in Table~\ref{tab:sim_self_aug_ablation}, ``Self w/o Aug'' performs the worst, confirming that directly reusing the exemplar pair causes severe spatial shortcut learning. Qualitative failure cases are provided in Appendix~\ref{app:shortcut}. In contrast, Self-Augmentation achieves comparable performance to Cross-Identity. This shows that using the same-identity self-augmented quadruplets during training does not harm cross-identity operation transfer.

\textbf{Self-Augmentation is more advantageous when operation alignment is difficult.}
We conduct an ablation on the professional retouching subset and evaluate on ProPortrait-500.
Compared with the simulated subset, the professional subset involves more diverse identities, shot scales, poses, portrait compositions, and locally applicable operations, making cross-identity training data much harder to keep operation-aligned.
We therefore compare the two relevant data construction strategies: Cross-Identity and Self-Augmentation.

\begin{table}[t]
\centering
\caption{Ablation results on ProPortrait-500 for evaluating the benefit of self-augmentation when cross-identity training data is difficult to align.}
\label{tab:pro_self_aug_ablation}
\begin{tabular}{lcccc}
\toprule
Setting & PSNR $\uparrow$ & SSIM $\uparrow$ & LPIPS $\downarrow$ & Face Similarity $\uparrow$ \\
\midrule
Cross-Identity & 30.52 & 0.914 & 0.212 & 0.916 \\
Self-Augmentation (Ours) & \textbf{32.65} & \textbf{0.927} & \textbf{0.200} & \textbf{0.960} \\
\bottomrule
\end{tabular}
\end{table}

As shown in Table~\ref{tab:pro_self_aug_ablation}, Self-Augmentation clearly outperforms Cross-Identity across all metrics. These results demonstrate that self-augmentation provides a more reliable training paradigm when cross-identity training data suffers from imperfect operation alignment.

We also compare the training efficiency between Self-Augmentation and Cross-Identity. As shown in Figure~\ref{fig:ablation_curve}, Self-Augmentation converges faster. This suggests that Self-Augmentation reduces the learning difficulty by preserving operation consistency while avoiding unnecessary interference from cross-identity pairing.

More ablation studies are provided in Appendix~\ref{app:ablation}.

\section{Conclusion}

In this paper, we introduce the novel task of Exemplar-Based Portrait Photo Retouching and propose MirrorPPR, which integrates a Retouching Operation Extractor with a Diffusion Transformer. To address operation misalignment and data scarcity, we develop a data Self-Augmentation paradigm and construct MirrorPPR47M, a large-scale dataset with over 47 million pairs for progressive curriculum learning. Extensive experiments demonstrate that MirrorPPR achieves state-of-the-art performance in retouching quality and identity preservation, establishing a solid foundation for real-world structural portrait retouching.

\begin{acks}
This work was supported by Shanghai Key Technology R\&D Program ``New Generation of Information Technology'' (No. 25511103700), NSF of China (Nos. 62306176, 92470118), CCF-ALIMAMA TECH Kangaroo Fund (NO. CCF-ALIMAMA OF 2025010), and Ant Group.
\end{acks}

\setlength{\bibsep}{5pt}
\bibliography{main}

@String(CVPR  = {IEEE Conf. Comput. Vis. Pattern Recog.})

@String(ICLR  = {Int. Conf. Learn. Represent.})

@String(AAAI  = {AAAI})

@String(TOG   = {ACM Trans. Graph.})

@String(CVPR  = {CVPR})

@String(ICLR  = {ICLR})

@String(TOG   = {ACM TOG})

@misc{brack2024leditslimitlessimageediting,
      title={LEDITS++: Limitless Image Editing using Text-to-Image Models}, 
      author={Manuel Brack and Felix Friedrich and Katharina Kornmeier and Linoy Tsaban and Patrick Schramowski and Kristian Kersting and Apolinário Passos},
      year={2024},
      eprint={2311.16711},
      archivePrefix={arXiv},
      primaryClass={cs.CV},
      url={https://arxiv.org/abs/2311.16711}, 
}

@misc{wu2025qwenimagetechnicalreport,
      title={Qwen-Image Technical Report}, 
      author={Chenfei Wu and Jiahao Li and Jingren Zhou and Junyang Lin and Kaiyuan Gao and Kun Yan and Sheng-ming Yin and Shuai Bai and Xiao Xu and Yilei Chen and Yuxiang Chen and Zecheng Tang and Zekai Zhang and Zhengyi Wang and An Yang and Bowen Yu and Chen Cheng and Dayiheng Liu and Deqing Li and Hang Zhang and Hao Meng and Hu Wei and Jingyuan Ni and Kai Chen and Kuan Cao and Liang Peng and Lin Qu and Minggang Wu and Peng Wang and Shuting Yu and Tingkun Wen and Wensen Feng and Xiaoxiao Xu and Yi Wang and Yichang Zhang and Yongqiang Zhu and Yujia Wu and Yuxuan Cai and Zenan Liu},
      year={2025},
      eprint={2508.02324},
      archivePrefix={arXiv},
      primaryClass={cs.CV},
      url={https://arxiv.org/abs/2508.02324}, 
}

@article{LongCat-Image,
      title={LongCat-Image Technical Report},
      author={Meituan LongCat Team and  Hanghang Ma and Haoxian Tan and Jiale Huang and Junqiang Wu and Jun-Yan He and Lishuai Gao and Songlin Xiao and Xiaoming Wei and Xiaoqi Ma and Xunliang Cai and Yayong Guan and Jie Hu},
	    journal={arXiv preprint arXiv:2512.07584},
      year={2025}
}

@article{liu2025step1x-edit,
  title={Step1X-Edit: A Practical Framework for General Image Editing}, 
  author={Shiyu Liu and Yucheng Han and Peng Xing and Fukun Yin and Rui Wang and Wei Cheng and Jiaqi Liao and Yingming Wang and Honghao Fu and Chunrui Han and Guopeng Li and Yuang Peng and Quan Sun and Jingwei Wu and Yan Cai and Zheng Ge and Ranchen Ming and Lei Xia and Xianfang Zeng and Yibo Zhu and Binxing Jiao and Xiangyu Zhang and Gang Yu and Daxin Jiang},
  journal={arXiv preprint arXiv:2504.17761},
  year={2025}
}

@article{wu2025omnigen2,
  title={OmniGen2: Exploration to Advanced Multimodal Generation},
  author={Chenyuan Wu and Pengfei Zheng and Ruiran Yan and Shitao Xiao and Xin Luo and Yueze Wang and Wanli Li and Xiyan Jiang and Yexin Liu and Junjie Zhou and Ze Liu and Ziyi Xia and Chaofan Li and Haoge Deng and Jiahao Wang and Kun Luo and Bo Zhang and Defu Lian and Xinlong Wang and Zhongyuan Wang and Tiejun Huang and Zheng Liu},
  journal={arXiv preprint arXiv:2506.18871},
  year={2025}
}

@misc{couairon2022diffeditdiffusionbasedsemanticimage,
      title={DiffEdit: Diffusion-based semantic image editing with mask guidance}, 
      author={Guillaume Couairon and Jakob Verbeek and Holger Schwenk and Matthieu Cord},
      year={2022},
      eprint={2210.11427},
      archivePrefix={arXiv},
      primaryClass={cs.CV},
      url={https://arxiv.org/abs/2210.11427}, 
}

@misc{flux-2-2025,
    author={Black Forest Labs},
    title={{FLUX.2: Frontier Visual Intelligence}},
    year={2025},
    howpublished={\url{https://bfl.ai/blog/flux-2}},
    note = {Accessed on June 24, 2026}
}

@article{seedream2025seedream,
  title={Seedream 4.0: Toward next-generation multimodal image generation},
  author={Seedream, Team and Chen, Yunpeng and Gao, Yu and Gong, Lixue and Guo, Meng and Guo, Qiushan and Guo, Zhiyao and Hou, Xiaoxia and Huang, Weilin and Huang, Yixuan and others},
  journal={arXiv preprint arXiv:2509.20427},
  year={2025}
}

@article{hertz2022prompt,
  title={Prompt-to-prompt image editing with cross attention control},
  author={Hertz, Amir and Mokady, Ron and Tenenbaum, Jay and Aberman, Kfir and Pritch, Yael and Cohen-Or, Daniel},
  journal={arXiv preprint arXiv:2208.01626},
  year={2022}
}

@inproceedings{brooks2023instructpix2pix,
  title={Instructpix2pix: Learning to follow image editing instructions},
  author={Brooks, Tim and Holynski, Aleksander and Efros, Alexei A},
  booktitle={Proceedings of the IEEE/CVF conference on computer vision and pattern recognition},
  pages={18392--18402},
  year={2023}
}

@article{meng2021sdedit,
  title={Sdedit: Guided image synthesis and editing with stochastic differential equations},
  author={Meng, Chenlin and He, Yutong and Song, Yang and Song, Jiaming and Wu, Jiajun and Zhu, Jun-Yan and Ermon, Stefano},
  journal={arXiv preprint arXiv:2108.01073},
  year={2021}
}

@inproceedings{sohl2015deep,
  title={Deep unsupervised learning using nonequilibrium thermodynamics},
  author={Sohl-Dickstein, Jascha and Weiss, Eric and Maheswaranathan, Niru and Ganguli, Surya},
  booktitle={International conference on machine learning},
  pages={2256--2265},
  year={2015},
  organization={pmlr}
}

@article{song2019generative,
  title={Generative modeling by estimating gradients of the data distribution},
  author={Song, Yang and Ermon, Stefano},
  journal={Advances in neural information processing systems},
  volume={32},
  year={2019}
}

@article{ho2020denoising,
  title={Denoising diffusion probabilistic models},
  author={Ho, Jonathan and Jain, Ajay and Abbeel, Pieter},
  journal={Advances in neural information processing systems},
  volume={33},
  pages={6840--6851},
  year={2020}
}

@inproceedings{cao2023masactrl,
  title={Masactrl: Tuning-free mutual self-attention control for consistent image synthesis and editing},
  author={Cao, Mingdeng and Wang, Xintao and Qi, Zhongang and Shan, Ying and Qie, Xiaohu and Zheng, Yinqiang},
  booktitle={Proceedings of the IEEE/CVF international conference on computer vision},
  pages={22560--22570},
  year={2023}
}

@inproceedings{sheynin2024emu,
  title={Emu edit: Precise image editing via recognition and generation tasks},
  author={Sheynin, Shelly and Polyak, Adam and Singer, Uriel and Kirstain, Yuval and Zohar, Amit and Ashual, Oron and Parikh, Devi and Taigman, Yaniv},
  booktitle={Proceedings of the IEEE/CVF Conference on Computer Vision and Pattern Recognition},
  pages={8871--8879},
  year={2024}
}

@article{hurst2024gpt,
  title={Gpt-4o system card},
  author={Hurst, Aaron and Lerer, Adam and Goucher, Adam P and Perelman, Adam and Ramesh, Aditya and Clark, Aidan and Ostrow, AJ and Welihinda, Akila and Hayes, Alan and Radford, Alec and others},
  journal={arXiv preprint arXiv:2410.21276},
  year={2024}
}

@article{li2023instructany2pix,
  title={Instructany2pix: Flexible visual editing via multimodal instruction following},
  author={Li, Shufan and Singh, Harkanwar and Grover, Aditya},
  journal={arXiv preprint arXiv:2312.06738},
  year={2023}
}

@misc{cai2026idglowdynamicidentitymodulation,
      title={IdGlow: Dynamic Identity Modulation for Multi-Subject Generation}, 
      author={Honghao Cai and Xiangyuan Wang and Yunhao Bai and Tianze Zhou and Sijie Xu and Yuyang Hao and Zezhou Cui and Yuyuan Yang and Wei Zhu and Yibo Chen and Xu Tang and Yao Hu and Zhen Li},
      year={2026},
      eprint={2603.00607},
      archivePrefix={arXiv},
      primaryClass={cs.CV},
      url={https://arxiv.org/abs/2603.00607}, 
}

@article{liao2017visual,
  title={Visual attribute transfer through deep image analogy},
  author={Liao, Jing and Yao, Yuan and Yuan, Lu and Hua, Gang and Kang, Sing Bing},
  journal={arXiv preprint arXiv:1705.01088},
  year={2017}
}

@article{gu2024analogist,
  title={Analogist: Out-of-the-box visual in-context learning with image diffusion model},
  author={Gu, Zheng and Yang, Shiyuan and Liao, Jing and Huo, Jing and Gao, Yang},
  journal={ACM Transactions on Graphics (TOG)},
  volume={43},
  number={4},
  pages={1--15},
  year={2024},
  publisher={ACM New York, NY, USA}
}

@inproceedings{yang2023paint,
  title={Paint by example: Exemplar-based image editing with diffusion models},
  author={Yang, Binxin and Gu, Shuyang and Zhang, Bo and Zhang, Ting and Chen, Xuejin and Sun, Xiaoyan and Chen, Dong and Wen, Fang},
  booktitle={Proceedings of the IEEE/CVF conference on computer vision and pattern recognition},
  pages={18381--18391},
  year={2023}
}

@misc{zhaoInstructBrushLearningAttentionbased2024a,
  title = {{{InstructBrush}}: {{Learning Attention-based Instruction Optimization}} for {{Image Editing}}},
  author = {Zhao, Ruoyu and Fan, Qingnan and Kou, Fei and Qin, Shuai and Gu, Hong and Wu, Wei and Xu, Pengcheng and Zhu, Mingrui and Wang, Nannan and Gao, Xinbo},
  year = 2024,
  number = {arXiv:2403.18660},
  eprint = {2403.18660},
  publisher = {arXiv},
  doi = {10.48550/arXiv.2403.18660},
}

@article{li2026viral,
  title={VIRAL: Visual In-Context Reasoning via Analogy in Diffusion Transformers},
  author={Li, Zhiwen and Duan, Zhongjie and Ye, Jinyan and Chen, Cen and Chen, Daoyuan and Li, Yaliang and Chen, Yingda},
  journal={arXiv preprint arXiv:2602.03210},
  year={2026}
}

@article{chen2025edit,
  title={Edit Transfer: Learning Image Editing via Vision In-Context Relations},
  author={Chen, Lan and Mao, Qi and Gu, Yuchao and Shou, Mike Zheng},
  journal={arXiv preprint arXiv:2503.13327},
  year={2025}
}

@article{gong2025relationadapter,
  title={Relationadapter: Learning and transferring visual relation with diffusion transformers},
  author={Gong, Yan and Song, Yiren and Li, Yicheng and Li, Chenglin and Zhang, Yin},
  journal={arXiv preprint arXiv:2506.02528},
  year={2025}
}

@inproceedings{xu2025textualize,
  title={Textualize visual prompt for image editing via diffusion bridge},
  author={Xu, Pengcheng and Fan, Qingnan and Kou, Fei and Qin, Shuai and Gu, Hong and Zhao, Ruoyu and Ling, Charles and Wang, Boyu},
  booktitle={Proceedings of the AAAI Conference on Artificial Intelligence},
  volume={39},
  number={20},
  pages={21779--21787},
  year={2025}
}

@inproceedings{lai2025unleashing,
  title={Unleashing in-context learning of autoregressive models for few-shot image manipulation},
  author={Lai, Bolin and Juefei-Xu, Felix and Liu, Miao and Dai, Xiaoliang and Mehta, Nikhil and Zhu, Chenguang and Huang, Zeyi and Rehg, James M and Lee, Sangmin and Zhang, Ning and others},
  booktitle={Proceedings of the IEEE/CVF Conference on Computer Vision and Pattern Recognition},
  pages={18346--18357},
  year={2025}
}

@article{srivastava2024reedit,
  title={Reedit: Multimodal exemplar-based image editing with diffusion models},
  author={Srivastava, Ashutosh and Menta, Tarun Ram and Java, Abhinav and Jadhav, Avadhoot and Singh, Silky and Jandial, Surgan and Krishnamurthy, Balaji},
  journal={arXiv preprint arXiv:2411.03982},
  year={2024}
}

@article{yang2023imagebrush,
  title={Imagebrush: Learning visual in-context instructions for exemplar-based image manipulation},
  author={Yang, Yifan and Peng, Houwen and Shen, Yifei and Yang, Yuqing and Hu, Han and Qiu, Lili and Koike, Hideki and others},
  journal={Advances in Neural Information Processing Systems},
  volume={36},
  pages={48723--48743},
  year={2023}
}

@article{nguyen2023visual,
  title={Visual instruction inversion: Image editing via image prompting},
  author={Nguyen, Thao and Li, Yuheng and Ojha, Utkarsh and Lee, Yong Jae},
  journal={Advances in Neural Information Processing Systems},
  volume={36},
  pages={9598--9613},
  year={2023}
}

@article{lu2025pairedit,
  title={PairEdit: Learning Semantic Variations for Exemplar-based Image Editing},
  author={Lu, Haoguang and Chen, Jiacheng and Yang, Zhenguo and Gnanha, Aurele Tohokantche and Wang, Fu Lee and Qing, Li and Mao, Xudong},
  journal={arXiv preprint arXiv:2506.07992},
  year={2025}
}

@article{wang2023context,
  title={In-context learning unlocked for diffusion models},
  author={Wang, Zhendong and Jiang, Yifan and Lu, Yadong and He, Pengcheng and Chen, Weizhu and Wang, Zhangyang and Zhou, Mingyuan and others},
  journal={Advances in Neural Information Processing Systems},
  volume={36},
  pages={8542--8562},
  year={2023}
}

@article{bharati2016detecting,
  title={Detecting facial retouching using supervised deep learning},
  author={Bharati, Aparna and Singh, Richa and Vatsa, Mayank and Bowyer, Kevin W},
  journal={IEEE Transactions on Information Forensics and Security},
  volume={11},
  number={9},
  pages={1903--1913},
  year={2016},
  publisher={IEEE}
}

@inproceedings{bharati2017demography,
  title={Demography-based facial retouching detection using subclass supervised sparse autoencoder},
  author={Bharati, Aparna and Vatsa, Mayank and Singh, Richa and Bowyer, Kevin W and Tong, Xin},
  booktitle={2017 IEEE international joint conference on biometrics (IJCB)},
  pages={474--482},
  year={2017},
  organization={IEEE}
}

@article{rathgeb2020prnu,
  title={PRNU-based detection of facial retouching},
  author={Rathgeb, Christian and Botaljov, Angelika and Stockhardt, Fabian and Isadskiy, Sergey and Debiasi, Luca and Uhl, Andreas and Busch, Christoph},
  journal={IET Biometrics},
  volume={9},
  number={4},
  pages={154--164},
  year={2020},
  publisher={Wiley Online Library}
}

@article{rathgeb2020differential,
  title={Differential detection of facial retouching: A multi-biometric approach},
  author={Rathgeb, Christian and Satnoianu, C-I and Haryanto, Nathania E and Bernardo, Kevin and Busch, Christoph},
  journal={IEEE Access},
  volume={8},
  pages={106373--106385},
  year={2020},
  publisher={IEEE}
}

@inproceedings{shafaei2021autoretouch,
  title={Autoretouch: Automatic professional face retouching},
  author={Shafaei, Alireza and Little, James J and Schmidt, Mark},
  booktitle={Proceedings of the IEEE/CVF Winter Conference on Applications of Computer Vision},
  pages={990--998},
  year={2021}
}

@inproceedings{bychkovsky2011learning,
  title={Learning photographic global tonal adjustment with a database of input/output image pairs},
  author={Bychkovsky, Vladimir and Paris, Sylvain and Chan, Eric and Durand, Fr{\'e}do},
  booktitle={CVPR 2011},
  pages={97--104},
  year={2011},
  organization={IEEE}
}

@article{cai2018learning,
  title={Learning a deep single image contrast enhancer from multi-exposure images},
  author={Cai, Jianrui and Gu, Shuhang and Zhang, Lei},
  journal={IEEE transactions on image processing},
  volume={27},
  number={4},
  pages={2049--2062},
  year={2018},
  publisher={IEEE}
}

@article{hu2018exposure,
  title={Exposure: A white-box photo post-processing framework},
  author={Hu, Yuanming and He, Hao and Xu, Chenxi and Wang, Baoyuan and Lin, Stephen},
  journal={ACM Transactions on Graphics (TOG)},
  volume={37},
  number={2},
  pages={1--17},
  year={2018},
  publisher={ACM New York, NY, USA}
}

@article{kim1997contrast,
  title={Contrast enhancement using brightness preserving bi-histogram equalization},
  author={Kim, Yeong-Taeg},
  journal={IEEE transactions on Consumer Electronics},
  volume={43},
  number={1},
  pages={1--8},
  year={1997},
  publisher={IEEE}
}

@inproceedings{li2018beautygan,
  title={Beautygan: Instance-level facial makeup transfer with deep generative adversarial network},
  author={Li, Tingting and Qian, Ruihe and Dong, Chao and Liu, Si and Yan, Qiong and Zhu, Wenwu and Lin, Liang},
  booktitle={Proceedings of the 26th ACM international conference on Multimedia},
  pages={645--653},
  year={2018}
}

@inproceedings{kosugi2020unpaired,
  title={Unpaired image enhancement featuring reinforcement-learning-controlled image editing software},
  author={Kosugi, Satoshi and Yamasaki, Toshihiko},
  booktitle={Proceedings of the AAAI conference on artificial intelligence},
  volume={34},
  number={07},
  pages={11296--11303},
  year={2020}
}

@incollection{mantiuk2008display,
  title={Display adaptive tone mapping},
  author={Mantiuk, Rafa{\l} and Daly, Scott and Kerofsky, Louis},
  booktitle={ACM SIGGRAPH 2008 papers},
  pages={1--10},
  year={2008}
}

@article{mukherjee2008enhancement,
  title={Enhancement of color images by scaling the DCT coefficients},
  author={Mukherjee, Jayanta and Mitra, Sanjit K},
  journal={IEEE Transactions on Image processing},
  volume={17},
  number={10},
  pages={1783--1794},
  year={2008},
  publisher={IEEE}
}

@inproceedings{karras2019style,
  title={A style-based generator architecture for generative adversarial networks},
  author={Karras, Tero and Laine, Samuli and Aila, Timo},
  booktitle={Proceedings of the IEEE/CVF conference on computer vision and pattern recognition},
  pages={4401--4410},
  year={2019}
}

@inproceedings{tewari2020stylerig,
  title={Stylerig: Rigging stylegan for 3d control over portrait images},
  author={Tewari, Ayush and Elgharib, Mohamed and Bharaj, Gaurav and Bernard, Florian and Seidel, Hans-Peter and P{\'e}rez, Patrick and Zollhofer, Michael and Theobalt, Christian},
  booktitle={Proceedings of the IEEE/CVF conference on computer vision and pattern recognition},
  pages={6142--6151},
  year={2020}
}

@inproceedings{medin2022most,
  title={MOST-GAN: 3D morphable StyleGAN for disentangled face image manipulation},
  author={Medin, Safa C and Egger, Bernhard and Cherian, Anoop and Wang, Ye and Tenenbaum, Joshua B and Liu, Xiaoming and Marks, Tim K},
  booktitle={Proceedings of the AAAI conference on artificial intelligence},
  volume={36},
  number={2},
  pages={1962--1971},
  year={2022}
}

@article{hu2022lora,
  title={Lora: Low-rank adaptation of large language models.},
  author={Hu, Edward J and Shen, Yelong and Wallis, Phillip and Allen-Zhu, Zeyuan and Li, Yuanzhi and Wang, Shean and Wang, Liang and Chen, Weizhu and others},
  journal={Iclr},
  volume={1},
  number={2},
  pages={3},
  year={2022}
}

@inproceedings{chen2025moto,
  title={Moto: Latent motion token as the bridging language for learning robot manipulation from videos},
  author={Chen, Yi and Ge, Yuying and Tang, Weiliang and Li, Yizhuo and Ge, Yixiao and Ding, Mingyu and Shan, Ying and Liu, Xihui},
  booktitle={Proceedings of the IEEE/CVF International Conference on Computer Vision},
  pages={19752--19763},
  year={2025}
}

@inproceedings{he2022masked,
  title={Masked autoencoders are scalable vision learners},
  author={He, Kaiming and Chen, Xinlei and Xie, Saining and Li, Yanghao and Doll{\'a}r, Piotr and Girshick, Ross},
  booktitle={Proceedings of the IEEE/CVF conference on computer vision and pattern recognition},
  pages={16000--16009},
  year={2022}
}

@inproceedings{zhang2018unreasonable,
  title={The unreasonable effectiveness of deep features as a perceptual metric},
  author={Zhang, Richard and Isola, Phillip and Efros, Alexei A and Shechtman, Eli and Wang, Oliver},
  booktitle={Proceedings of the IEEE conference on computer vision and pattern recognition},
  pages={586--595},
  year={2018}
}

@misc{bai2025qwen25vltechnicalreport,
      title={Qwen2.5-VL Technical Report}, 
      author={Shuai Bai and Keqin Chen and Xuejing Liu and Jialin Wang and Wenbin Ge and Sibo Song and Kai Dang and Peng Wang and Shijie Wang and Jun Tang and Humen Zhong and Yuanzhi Zhu and Mingkun Yang and Zhaohai Li and Jianqiang Wan and Pengfei Wang and Wei Ding and Zheren Fu and Yiheng Xu and Jiabo Ye and Xi Zhang and Tianbao Xie and Zesen Cheng and Hang Zhang and Zhibo Yang and Haiyang Xu and Junyang Lin},
      year={2025},
      eprint={2502.13923},
      archivePrefix={arXiv},
      primaryClass={cs.CV},
      url={https://arxiv.org/abs/2502.13923}, 
}

@inproceedings{liang2021ppr10k,
  title={Ppr10k: A large-scale portrait photo retouching dataset with human-region mask and group-level consistency},
  author={Liang, Jie and Zeng, Hui and Cui, Miaomiao and Xie, Xuansong and Zhang, Lei},
  booktitle={Proceedings of the IEEE/CVF Conference on Computer Vision and Pattern Recognition},
  pages={653--661},
  year={2021}
}

@inproceedings{liu2025mofrr,
  title={MoFRR: Mixture of Diffusion Models for Face Retouching Restoration},
  author={Liu, Jiaxin and Ying, Qichao and Qian, Zhenxing and Li, Sheng and Zhang, Runqi and Liu, Jian and Zhang, Xinpeng},
  booktitle={Proceedings of the IEEE/CVF International Conference on Computer Vision},
  pages={12842--12851},
  year={2025}
}

@inproceedings{deng2019arcface,
  title={Arcface: Additive angular margin loss for deep face recognition},
  author={Deng, Jiankang and Guo, Jia and Xue, Niannan and Zafeiriou, Stefanos},
  booktitle={Proceedings of the IEEE/CVF conference on computer vision and pattern recognition},
  pages={4690--4699},
  year={2019}
}

@article{pan2025transfer,
  title={Transfer between modalities with metaqueries},
  author={Pan, Xichen and Shukla, Satya Narayan and Singh, Aashu and Zhao, Zhuokai and Mishra, Shlok Kumar and Wang, Jialiang and Xu, Zhiyang and Chen, Jiuhai and Li, Kunpeng and Juefei-Xu, Felix and others},
  journal={arXiv preprint arXiv:2504.06256},
  year={2025}
}

@article{team2023gemini,
  title={Gemini: a family of highly capable multimodal models},
  author={Team, Gemini and Anil, Rohan and Borgeaud, Sebastian and Alayrac, Jean-Baptiste and Yu, Jiahui and Soricut, Radu and Schalkwyk, Johan and Dai, Andrew M and Hauth, Anja and Millican, Katie and others},
  journal={arXiv preprint arXiv:2312.11805},
  year={2023}
}

@inproceedings{bengio2009curriculum,
  title={Curriculum learning},
  author={Bengio, Yoshua and Louradour, J{\'e}r{\^o}me and Collobert, Ronan and Weston, Jason},
  booktitle={Proceedings of the 26th annual international conference on machine learning},
  pages={41--48},
  year={2009}
}

@article{wang2004image,
  title={Image quality assessment: from error visibility to structural similarity},
  author={Wang, Zhou and Bovik, Alan C and Sheikh, Hamid R and Simoncelli, Eero P},
  journal={IEEE transactions on image processing},
  volume={13},
  number={4},
  pages={600--612},
  year={2004},
  publisher={IEEE}
}

@misc{qwen_image_edit_2024,
  author       = {{DiffSynth-Studio}},
  title        = {Qwen-Image-Edit-2511-ICEdit-LoRA},
  howpublished = {Hugging Face Model Repository},
  year         = {2025},
  url          = {https://huggingface.co/DiffSynth-Studio/Qwen-Image-Edit-2511-ICEdit-LoRA},
  note = {Accessed on June 24, 2026}
}

@misc{dcgm_ffhq_features,
  author       = {DCGM},
  title        = {ffhq-features-dataset: Gender, Age, and Emotion for Flickr-Faces-HQ Dataset (FFHQ)},
  year         = {2019},
  publisher    = {GitHub},
  journal      = {GitHub repository},
  howpublished = {\url{https://github.com/DCGM/ffhq-features-dataset}},
  note = {Accessed on June 24, 2026}
}

@software{yolo11_ultralytics,
  author = {Glenn Jocher and Jing Qiu},
  title = {Ultralytics YOLO11},
  version = {11.0.0},
  year = {2024},
  url = {https://github.com/ultralytics/ultralytics},
  orcid = {0000-0001-5950-6979, 0000-0003-3783-7069},
  license = {AGPL-3.0},
  note = {Accessed on June 24, 2026}
}

@incollection{schaefer2006image,
  title={Image deformation using moving least squares},
  author={Schaefer, Scott and McPhail, Travis and Warren, Joe},
  booktitle={ACM SIGGRAPH 2006 Papers},
  pages={533--540},
  year={2006}
}

@article{lugaresi2019mediapipe,
  title={Mediapipe: A framework for building perception pipelines},
  author={Lugaresi, Camillo and Tang, Jiuqiang and Nash, Hadon and McClanahan, Chris and Uboweja, Esha and Hays, Michael and Zhang, Fan and Chang, Chuo-Ling and Yong, Ming Guang and Lee, Juhyun and others},
  journal={arXiv preprint arXiv:1906.08172},
  year={2019}
}

@misc{qwen2025qwen25technicalreport,
      title={Qwen2.5 Technical Report}, 
      author={{Qwen Team}},
      year={2025},
      eprint={2412.15115},
      archivePrefix={arXiv},
      primaryClass={cs.CL},
      url={https://arxiv.org/abs/2412.15115}, 
}

@inproceedings{xie2023revealing,
  title={Revealing the dark secrets of masked image modeling},
  author={Xie, Zhenda and Geng, Zigang and Hu, Jingcheng and Zhang, Zheng and Hu, Han and Cao, Yue},
  booktitle={Proceedings of the IEEE/CVF conference on computer vision and pattern recognition},
  pages={14475--14485},
  year={2023}
}
\bibliographystyle{splncs04}

\clearpage
\appendix
\renewcommand*{\theHsection}{appendix.\Alph{section}}
\renewcommand*{\theHsubsection}{\theHsection.\arabic{subsection}}
\renewcommand*{\theHsubsubsection}{\theHsubsection.\arabic{subsubsection}}
\renewcommand*{\theHequation}{\theHsection.\arabic{equation}}
\section{Dataset Details}
\label{sec:dataset_details}

In this section, we provide comprehensive details regarding the construction of the \textbf{MirrorPPR47M} dataset. This includes the data filtering criteria, the principles and overall pipeline of our Landmark-Guided Local Warping (LLW) algorithm, the taxonomy and statistics of retouching operations, and the data self-augmentation pipeline.

\subsection{Data Filtering Criteria}
\label{app:Data Filtering Criteria}
To ensure the high quality of the training data, we applied strict filtering criteria to the raw images before executing any retouching operations. 

\noindent \textbf{Simulated Retouching Subset.} The raw images for this subset are sourced from the FFHQ dataset~\cite{karras2019style}. We utilize comprehensive attribute annotations provided by the ffhq-features-dataset~\cite{dcgm_ffhq_features} to filter out low-quality samples. Specifically, we retain only images where the head pose angles (pitch, roll, and yaw) are all within $\pm 15^\circ$. Furthermore, we discard images that exhibit severe blurriness, poor exposure, high noise levels, or significant occlusions over critical facial regions.

\noindent \textbf{Professional Retouching Subset.} The raw high-resolution images are sourced from the PPR10K dataset~\cite{liang2021ppr10k}. We utilize YOLO~\cite{yolo11_ultralytics} to obtain the bounding boxes and person count. To guarantee that the network focuses on single-subject portrait retouching, we strictly filter out images containing more than one person. Additionally, to ensure sufficient resolution for capturing subtle retouching operations, we remove any images where the detected portrait area is less than $240,000$ pixels.

\subsection{Landmark-Guided Local Warping (LLW) Algorithm}
\label{app:LLW}
To construct the Simulated Retouching Subset, we propose the Landmark-Guided Local Warping (LLW) algorithm, which synthesizes paired images with relatively distinct modifications. The algorithm leverages facial landmarks to guide the Moving Least Squares (MLS) deformation~\cite{schaefer2006image}. The overall pipeline proceeds as follows:

\noindent \textbf{Landmark Detection and Control Point Definition.} 
Given a source image, we first extract 468 dense 2D facial landmarks using MediaPipe~\cite{lugaresi2019mediapipe}. Depending on the specified retouching operation (e.g., eye resizing, nose length adjustment), we define a set of \textit{moving points} (points to be displaced) and a set of \textit{anchor points} (points to remain fixed to prevent global distortion). Collectively, the moving points and anchor points constitute the \textit{control points}. Let the original coordinates of these control points be the source point set $\mathbf{P} = \{\mathbf{p}_i\}_{i=1}^N$. Based on manually defined deformation rules for each retouching operation, we compute their corresponding target coordinates $\mathbf{Q} = \{\mathbf{q}_i\}_{i=1}^N$.

\noindent \textbf{Similarity Moving Least Squares (MLS).} 
To smoothly warp the image while preserving the local structure, we employ Similarity MLS. In practice, for any pixel coordinate $\mathbf{v}$ in the target grid, we calculate its corresponding source coordinate $f(\mathbf{v})$ based on the control point mapping $\mathbf{q}_i \to \mathbf{p}_i$. 

First, we compute the spatially varying weights $w_i$ for each control point:
\begin{equation}
    w_i(\mathbf{v}) = \frac{1}{\|\mathbf{v} - \mathbf{q}_i\|^{2\alpha} + \epsilon},
\end{equation}
where $\alpha=1.0$ is the distance attenuation factor and $\epsilon=10^{-8}$ prevents zero division. We then calculate the weighted centroids of the target and source points:
\begin{equation}
    \mathbf{q}^* = \frac{\sum_i w_i \mathbf{q}_i}{\sum_i w_i}, \quad \mathbf{p}^* = \frac{\sum_i w_i \mathbf{p}_i}{\sum_i w_i}.
\end{equation}
Next, we obtain the centered coordinates $\hat{\mathbf{q}}_i = \mathbf{q}_i - \mathbf{q}^*$ and $\hat{\mathbf{p}}_i = \mathbf{p}_i - \mathbf{p}^*$. For a 2D vector $\mathbf{u} = (x, y)^\top$, we define its perpendicular vector as $\mathbf{u}^\perp = (-y, x)^\top$. Under the similarity transformation, the mapped source coordinate $f(\mathbf{v})$ is formulated as:
\begin{equation}
    f(\mathbf{v}) = \mathbf{p}^* + \frac{1}{\mu_s} \sum_{i=1}^N w_i \left[ \left( (\mathbf{v} - \mathbf{q}^*) \cdot \hat{\mathbf{q}}_i \right) \hat{\mathbf{p}}_i + \left( (\mathbf{v} - \mathbf{q}^*) \cdot \hat{\mathbf{q}}_i^\perp \right) \hat{\mathbf{p}}_i^\perp \right],
\end{equation}
where $\mu_s = \sum_i w_i \|\hat{\mathbf{q}}_i\|^2$. Finally, we sample the pixel values at $f(\mathbf{v})$ using bilinear interpolation to generate the warped image $I_{warp}$.

\noindent \textbf{Mask-based Feathering and Blending.} 
To ensure seamless integration of the retouched region with the unedited background, we compute the convex hull of the affected landmarks to generate a binary mask. A Gaussian blur is applied to the mask to create a soft feathering effect $M \in [0, 1]$. The final retouched image $I_{out}$ is obtained by alpha blending:
\begin{equation}
    I_{out} = M \odot I_{warp} + (1 - M) \odot I_{orig},
\end{equation}
where $I_{orig}$ denotes the original unedited image, and $\odot$ denotes element-wise multiplication.

\subsection{Data Processing and Self-Augmentation}
\label{app:data process}
To adapt to the model architecture and improve training efficiency, we implement a carefully designed data processing pipeline. All images are standardized to a total area of approximately $4$ MP, with both height and width strictly constrained to be multiples of 16. 

For the simulated subset, the original images are directly resized to the target area. For the professional subset, we first obtain the bounding box of the portrait region detected by YOLO. We then expand this bounding box outwards by $10\%$ on all sides. The image is cropped according to this expanded bounding box and subsequently resized to the target area. These processed images serve as our base exemplar pairs.

Following this, we apply randomized data self-augmentation to both images in an exemplar pair. The augmentations include:
\begin{itemize}
    \item \textbf{Rotation:} Randomly applied at discrete intervals of $\pm 5^\circ, \pm 10^\circ$, and $\pm 15^\circ$.
    \item \textbf{Horizontal Flipping:} Randomly mirroring the images left-to-right.
    \item \textbf{Dynamic Cropping:} A single original image pair is randomly cropped into multiple versions with extreme aspect ratios ranging from 1:3 to 3:1. 
\end{itemize}
During the random cropping process, we leverage the facial bounding boxes provided by the FFHQ dataset and the human body masks detected by YOLO to ensure that the augmented views strictly retain the main portrait subject. Ultimately, we generate an average of $13.3$ augmented variations per pair in the simulated subset, and $13.4$ variations per pair in the professional subset.

\subsection{Taxonomy and Statistics of Retouching Operations}
\label{app:supp_operation_taxonomy_statistics}

To cover a comprehensive range of structural portrait retouching, MirrorPPR47M encompasses diverse operations over facial features, face shapes, and body proportions.

In the simulated subset, we implement 8 base operation types using the Landmark-Guided Local Warping (LLW) algorithm. Each type supports two opposite directions, resulting in 16 directed operations, as listed in Table~\ref{tab:simulated_ops}. The operation-type distribution is uniform: each of the 8 base operation types accounts for 12.5\%. The number of operations per exemplar pair is also uniformly distributed from 1 to 8, with each operation count accounting for 12.5\%.

The professional subset covers 27 fine-grained retouching operations, including 18 facial-feature operations, 4 face-shape operations, and 5 body-proportion operations, as detailed in Table~\ref{tab:professional_ops}. For the operation-type distribution, each of the 22 facial and face-shape operations accounts for approximately 3.4\% of all sampled operation instances, while each of the 5 body-proportion operations accounts for approximately 5.0\%. For the number of operations per exemplar pair, pairs containing 1--7 operations account for 0.2\%, 2.5\%, 18.2\%, 19.7\%, 20.1\%, 20.2\%, and 19.1\%, respectively. The relatively small proportions of one- and two-operation pairs arise from their limited combinatorial space over the 3,789 source portraits.

\begin{table}[t]
\centering
\caption{Taxonomy of the 16 retouching operations in the simulated subset.}
\label{tab:simulated_ops}
\renewcommand{\arraystretch}{1.3}
\begin{tabular}{l | >{\raggedright\arraybackslash}p{0.75\linewidth}}
\toprule
\textbf{Category} & \textbf{Operations (Bidirectional)} \\
\midrule
Eyes & Decrease/Increase eye distance, Shrink/Enlarge eyes \\
\midrule
Nose & Narrow/Widen nose bridge, Narrow/Widen nasal alae, Shorten/Lengthen nose \\
\midrule
Mouth & Move mouth downward/upward, Thin/Plump lips, Shrink/Enlarge mouth \\
\bottomrule
\end{tabular}
\end{table}

\begin{table}[t]
\centering
\caption{Taxonomy of the 27 fine-grained retouching operations in the professional subset.}
\label{tab:professional_ops}
\renewcommand{\arraystretch}{1.4}
\begin{tabular}{l | >{\raggedright\arraybackslash}p{0.75\linewidth}}
\toprule
\textbf{Category} & \textbf{Operations} \\
\midrule
Face Shape & Round/Sharpen face, Sharpen/Square jawline, Shorten/Lengthen chin, Depress/Fill temples \\
\midrule
Eyes \& Brows & Decrease/Increase eyebrow distance, Move eyebrows downward/upward, Thin/Thicken eyebrows, Decrease/Increase eye distance, Decrease/Increase eye height, Decrease/Increase eye width, Shrink/Enlarge eyes, Move eyes downward/upward \\
\midrule
Nose & Shrink/Enlarge nose, Narrow/Widen nasal alae, Narrow/Widen nose bridge, Shrink/Enlarge nose tip, Shorten/Lengthen nose \\
\midrule
Mouth & Move mouth downward/upward, Thin/Plump lips, Lower/Raise mouth corners, Narrow/Widen mouth, Shrink/Enlarge mouth \\
\midrule
Body & Square shoulders, Narrow/Broaden shoulders, Thicken/Slim arms, Slim legs, Slim waist \\
\bottomrule
\end{tabular}
\end{table}

\section{Model Architecture Details}
\label{app:arch_details}

In this section, we provide detailed configurations regarding the network architecture and input processing. The trainable components of our framework mainly consist of the R-Former, the auxiliary ViT decoder, the connector, and the LoRA modules. 

\noindent\textbf{Input Area.} 
The input to the Retouching Operation Extractor is resized to an area of approximately $2048 \times 2048$ pixels. The input to the Qwen2.5-VL~\cite{bai2025qwen25vltechnicalreport} module is resized to an area of approximately $512 \times 512$ pixels. Following the recommended settings of the Qwen-Image-Edit-2511~\cite{wu2025qwenimagetechnicalreport} backbone, the input to the VAE encoder is resized to an area of approximately $1536 \times 1536$ pixels.

\noindent\textbf{Module Designs.} 
Both the R-Former and the ViT decoder use the standard Vision Transformer (ViT) architecture. Specifically, the R-Former incorporates a set of internal learnable query tokens. The connector bridges the R-Former and the frozen DiT. Following MetaQuery~\cite{pan2025transfer}, it adopts the same architecture as the Qwen2.5 LLM~\cite{qwen2025qwen25technicalreport} while enabling bi-directional attention. It utilizes an Enc-Proj design: a Transformer encoder first aligns the features to the instruction conditioning space at the dimensionality of $\mathbf{H}_{edit}$; subsequently, a linear projection layer projects these conditions into the input dimension of the DiT blocks. 

Detailed hyperparameter configurations for these core modules are summarized in Table~\ref{tab:model_arch}.

\begin{table}[t]
\centering
\caption{\textbf{Detailed Model Architecture Hyperparameters.} This table outlines the specific configurations for the R-Former, the auxiliary ViT decoder, the connector, and the LoRA modules.}
\label{tab:model_arch}
\renewcommand{\arraystretch}{1.2}
\setlength{\tabcolsep}{12pt}
\begin{tabular}{llc}
\toprule
\textbf{Component} & \textbf{Parameter} & \textbf{Value} \\
\midrule
\multirow{4}{*}{R-Former} 
& \texttt{num\_queries} & 8 \\
& \texttt{num\_layers} & 4 \\
& \texttt{hidden\_size} & 768 \\
& \texttt{num\_heads} & 12 \\
\midrule
\multirow{4}{*}{ViT Decoder} 
& \texttt{patch\_size} & 16 \\
& \texttt{num\_layers} & 12 \\
& \texttt{hidden\_size} & 768 \\
& \texttt{num\_heads} & 12 \\
\midrule
\multirow{4}{*}{Connector} 
& \texttt{hidden\_size} & 768 \\
& \texttt{num\_layers} & 6 \\
& \texttt{num\_heads} & 12 \\
& \texttt{output\_dim} & 3584 \\
\midrule
\multirow{4}{*}{LoRA} 
& \texttt{lora\_rank} & 32 \\
& \texttt{lora\_alpha} & 32 \\
& \texttt{lora\_dropout} & 0.0 \\
& \texttt{bias} & none \\
\bottomrule
\end{tabular}
\end{table}

\section{Training Details}
\label{app:training_details}

We train two model variants, \textbf{MirrorPPR-Face} and \textbf{MirrorPPR-Pro}, with training steps tailored to their respective complexities:

\begin{itemize}
    \item \textbf{MirrorPPR-Face:} The Retouching Operation Extractor is first pre-trained for approximately $40,000$ steps using the auxiliary reconstruction task. Afterward, the ViT decoder is discarded, and the entire framework is jointly fine-tuned for $150,000$ steps.
    
    \item \textbf{MirrorPPR-Pro:} This model follows a progressive curriculum learning strategy. First, the extractor is pre-trained on the simulated subset for approximately $60,000$ steps. It is then further pre-trained on the professional subset for $40,000$ steps. Finally, the entire framework is jointly fine-tuned exclusively on the professional subset for $150,000$ steps.
\end{itemize}

The complete set of optimization hyperparameters, including learning rate, scheduler, and weight decay for both training stages, is detailed in Table~\ref{tab:training_settings}.

\begin{table}[t]
\centering
\caption{\textbf{Training Settings.} Summary of optimization hyperparameters for the progressive training pipeline.}
\label{tab:training_settings}
\renewcommand{\arraystretch}{1.2}
\setlength{\tabcolsep}{10pt}
\begin{tabular}{lll}
\toprule
\textbf{Stage} & \textbf{Parameter} & \textbf{Value} \\
\midrule
\multirow{7}{*}{Extractor Pre-training} 
& \texttt{batch\_size} & 1536 (simulated subset) \\
&                      & 512 (professional subset) \\
& \texttt{optimizer} & AdamW \\
& \texttt{lr\_max} & 1e-4 \\
& \texttt{lr\_min} & 5e-5 \\
& \texttt{lr\_scheduler} & cosine decay \\
& \texttt{weight\_decay} & 1e-4 \\
& \texttt{balancing\_scalar} & 1.0 \\
\midrule
\multirow{5}{*}{Joint Fine-tuning} 
& \texttt{batch\_size} & 64 \\
& \texttt{optimizer} & AdamW \\
& \texttt{lr} & 1e-5 \\
& \texttt{lr\_scheduler} & constant \\
& \texttt{weight\_decay} & 1e-2 \\
\bottomrule
\end{tabular}
\end{table}

\section{Evaluation Details}
\label{app:eval_details}

The inference configurations for all evaluated models are detailed in Table~\ref{tab:inference_details}. We adopt the officially recommended inference configurations for each model. By default, the output resolution matches the query image: SimFace-100 uses an area equivalent to 1024×1024 pixels, while ProPortrait-500 uses an area equivalent to $1.5\text{K} \times 1.5\text{K}$ pixels. These notations indicate approximate pixel areas rather than fixed square dimensions.
Due to specific resolution constraints, the maximum output areas for Nano Banana 2~\cite{team2023gemini}, Seedream 4.5~\cite{seedream2025seedream}, and EditTransfer~\cite{chen2025edit} are equivalent to $1024 \times 1024$, $2048 \times 2048$, and $512 \times 512$, respectively. Before calculating the evaluation metrics, all generated outputs are resized to the same dimensions as the query image.

\begin{table}[t]
\centering
\caption{Inference configurations for all evaluated models. ``default'' indicates that the output size follows the input query image. The query images have an area equivalent to 1024×1024 pixels on SimFace-100 and $1.5\text{K} \times 1.5\text{K}$ pixels on ProPortrait-500.}
\label{tab:inference_details}
\resizebox{\linewidth}{!}{
\begin{tabular}{ll c c@{\hspace{3em}}c}
\toprule
\multirow{2}{*}{\textbf{Category}} & \multirow{2}{*}{\textbf{Model}} & \textbf{Inference} & \multicolumn{2}{c}{\textbf{Output Area}} \\
\cmidrule(lr){4-5}
& & \textbf{Steps} & \textbf{SimFace-100} & \textbf{ProPortrait-500} \\
\midrule
\multirow{4}{*}{\begin{tabular}[c]{@{}l@{}}Multi-reference\\Image Editing\\/ Text-guided\end{tabular}} 
& Qwen-Image-Edit-2511             & 40 & default & default \\
& FLUX.2-dev                       & 50 & default & default \\
& Nano Banana 2                    & $-$  & default & $1024 \times 1024$ \\
& Seedream 4.5                     & $-$  & $2048 \times 2048$ & $2048 \times 2048$ \\
\midrule
\multirow{3}{*}{Exemplar-based} 
& Qwen-Image-Edit-2511-ICEdit-LoRA & 50 & default & default \\
& RelationAdapter                  & 24 & default & default \\
& EditTransfer                     & 35 & $512 \times 512$   & $512 \times 512$ \\
\midrule
Ours & MirrorPPR-Face / MirrorPPR-Pro & 40 & default & default \\
\bottomrule
\end{tabular}
}
\end{table}

\section{User Study}
\label{app:user_study}

\begin{figure}[t]
\centering
\includegraphics[width=\linewidth]{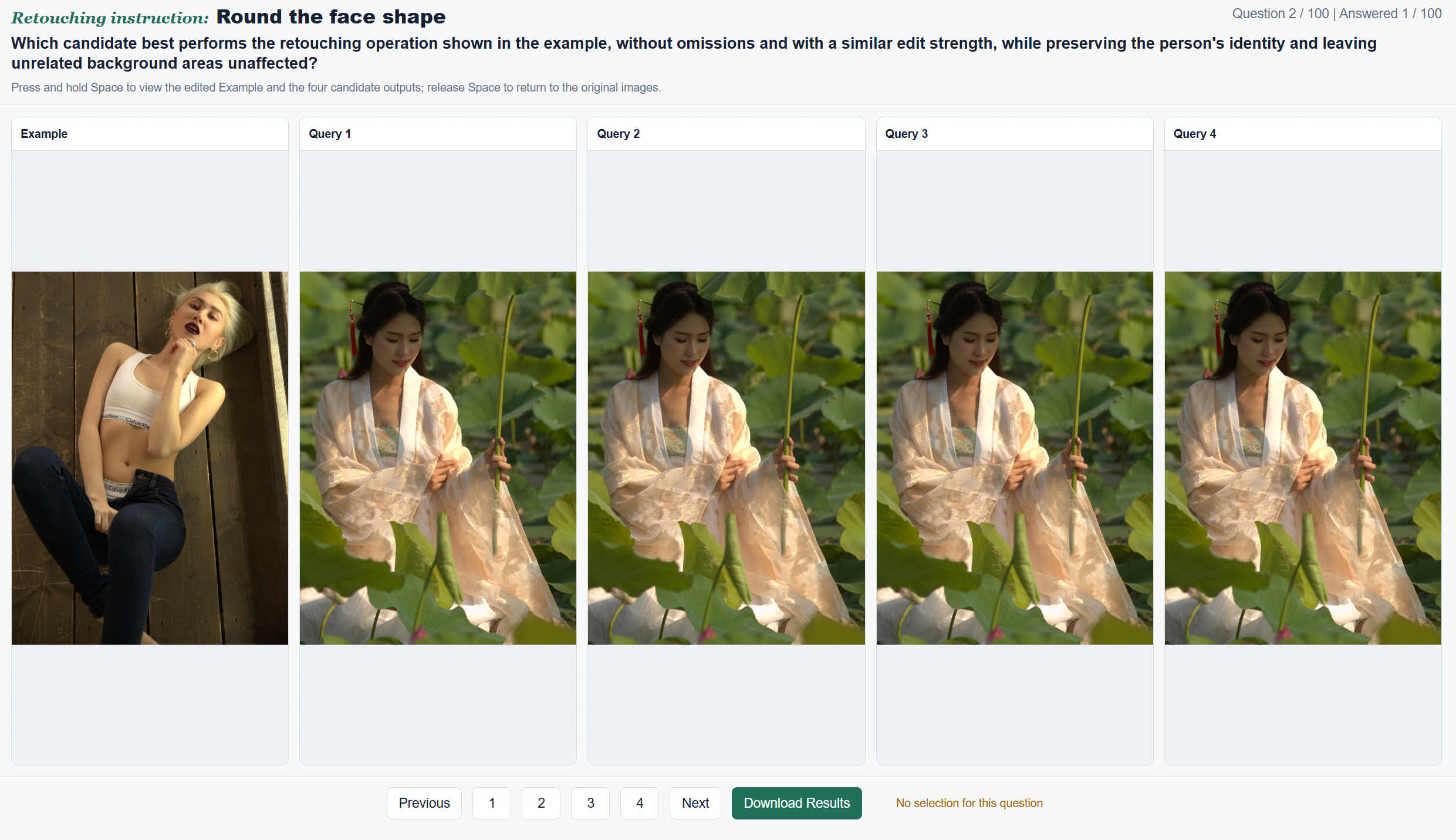}
\caption{Screenshot of the web interface used in the user study. Candidate outputs are anonymized as numbered options, and participants select the result that best transfers the demonstrated retouching operation while preserving identity and unrelated regions.}
\label{fig:user_study_interface}
\end{figure}

We conduct a user study on 100 samples from ProPortrait-500. We compare MirrorPPR-Pro with the strongest representative baseline from each evaluated category: Nano Banana 2 for multi-reference image editing, EditTransfer for exemplar-based editing, and Nano Banana 2 for text-guided editing. Each question presents the exemplar pair, the query image, and four anonymized outputs. Participants are asked to select the candidate that best performs the demonstrated retouching operation with a similar edit strength, while preserving identity and leaving unrelated background regions unaffected. We collect responses from 30 users and compute each method's preference as the percentage of total selections. The web interface is shown in Figure~\ref{fig:user_study_interface}, and the full results are reported in Table~\ref{tab:user_study}.

\begin{table}[t]
\centering
\caption{User study results on ProPortrait-500. Preference denotes the percentage of selections among all responses.}
\label{tab:user_study}
\begin{tabular}{llc}
\toprule
Category & Method & Preference (\%) $\uparrow$ \\
\midrule
Multi-reference Image Editing & Nano Banana 2 & 0.4 \\
Exemplar-based & EditTransfer & 0.0 \\
Text-guided & Nano Banana 2 & 20.6 \\
Ours & MirrorPPR-Pro & \textbf{79.0} \\
\bottomrule
\end{tabular}
\end{table}

\section{More Ablation Studies}
\label{app:ablation}

\subsection{Failure Analysis of the ``Self w/o Aug'' Strategy}
\label{app:shortcut}
\begin{figure}[t]
  \centering
  \includegraphics[width=\textwidth]{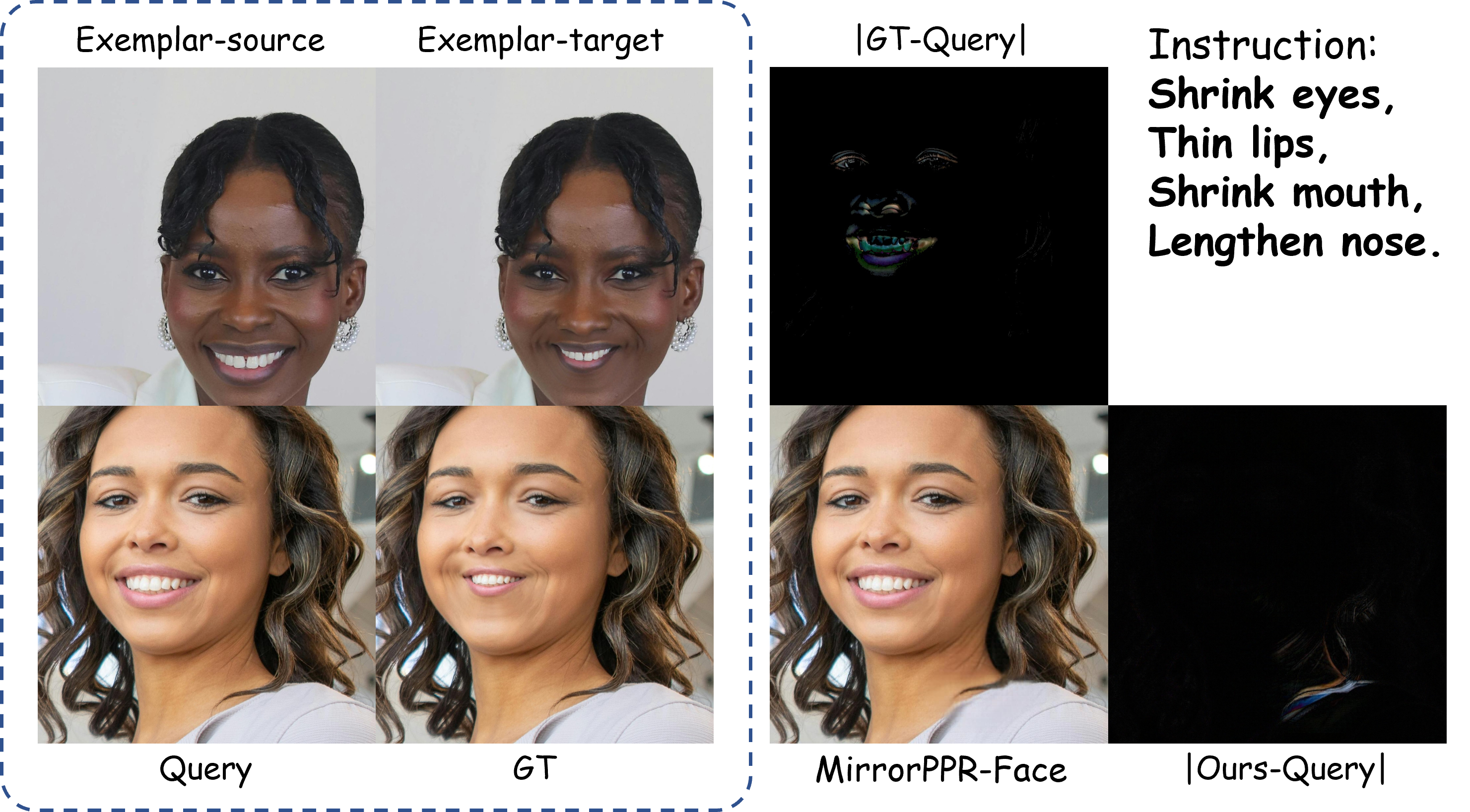}
  \caption{\textbf{Failure analysis of ``Self w/o Aug''.} The model overfits to the absolute edit position in the exemplar and fails to transfer the operations to the query image.}
  \label{fig:self_wo_aug}
\end{figure}

As mentioned in Section 5.4 of the main paper, we provide a qualitative failure case for the ``Self w/o Aug'' setting in Figure~\ref{fig:self_wo_aug}. As shown by the difference map, without spatial augmentation, the model memorizes the absolute spatial coordinates of the edits (bottom-right) and erroneously modifies the irrelevant background of the query image, failing to edit the face in the top-left of the query image. This clearly demonstrates severe spatial shortcut learning, further validating the necessity of spatial augmentations during training.

\subsection{Model Design Ablations}

We conduct additional ablation studies on key model design choices, including the number of learnable query tokens in the Retouching
Operation Extractor, the depth of the R-Former, and the LoRA rank used during
joint fine-tuning. All variants are trained under the MirrorPPR-Face setting and
evaluated on the SimFace-100 benchmark. For each group, only the specified
design choice is changed while all other settings are kept the same as the
default configuration.

As shown in Table~\ref{tab:network_design_ablation}, increasing the number of learnable query tokens brings only marginal metric gains. However, more query tokens increase the length of the operation condition injected into the diffusion backbone, increasing training and inference overhead. Therefore, we use $N_{\mathrm{query}}=8$ as the default setting.

For the R-Former depth, the 4-layer variant achieves the best performance. A shallower 2-layer R-Former slightly reduces reconstruction
quality, while increasing the depth to 8 layers does not bring further gains and
decreases Face Similarity. This suggests that a moderate depth is sufficient for
extracting subtle retouching operations, whereas deeper extractors may introduce
redundancy.

For the LoRA rank, $r=8$ limits the adaptation capacity and consistently
underperforms the default setting. Although $r=64$ slightly improves PSNR and
LPIPS, it reduces Face Similarity and introduces additional computational
overhead. We therefore choose $r=32$ as the default rank, which provides the
best trade-off 

\begin{table}[t]
\centering
\caption{Ablation study on key model design choices evaluated on SimFace-100. The default configuration uses
$N_{\mathrm{query}}=8$, a 4-layer R-Former, and LoRA rank $r=32$.}
\label{tab:network_design_ablation}
\begin{tabular}{lcccc}
\toprule
Setting & PSNR $\uparrow$ & SSIM $\uparrow$ & LPIPS $\downarrow$ & Face Similarity $\uparrow$ \\
\midrule
$N_{\mathrm{query}}=8$ (default) & 32.25 & 0.909 & \textbf{0.186} & 0.937 \\
$N_{\mathrm{query}}=64$ & 32.32 & \textbf{0.910} & \textbf{0.186} & 0.938 \\
$N_{\mathrm{query}}=256$ & \textbf{32.36} & 0.909 & 0.187 & \textbf{0.940} \\
\midrule
R-Former layers $=2$ & 32.10 & 0.908 & 0.189 & 0.934 \\
R-Former layers $=4$ (default) & \textbf{32.25} & \textbf{0.909} & \textbf{0.186} & \textbf{0.937} \\
R-Former layers $=8$ & 32.16 & 0.908 & 0.187 & 0.920 \\
\midrule
LoRA rank $r=8$ & 31.56 & 0.907 & 0.193 & 0.926 \\
LoRA rank $r=32$ (default) & 32.25 & \textbf{0.909} & 0.186 & \textbf{0.937} \\
LoRA rank $r=64$ & \textbf{32.29} & \textbf{0.909} & \textbf{0.184} & 0.933 \\
\bottomrule
\end{tabular}
\end{table}

\section{More Qualitative Examples}

\begin{figure}[t]
  \centering
    \includegraphics[width=\textwidth]{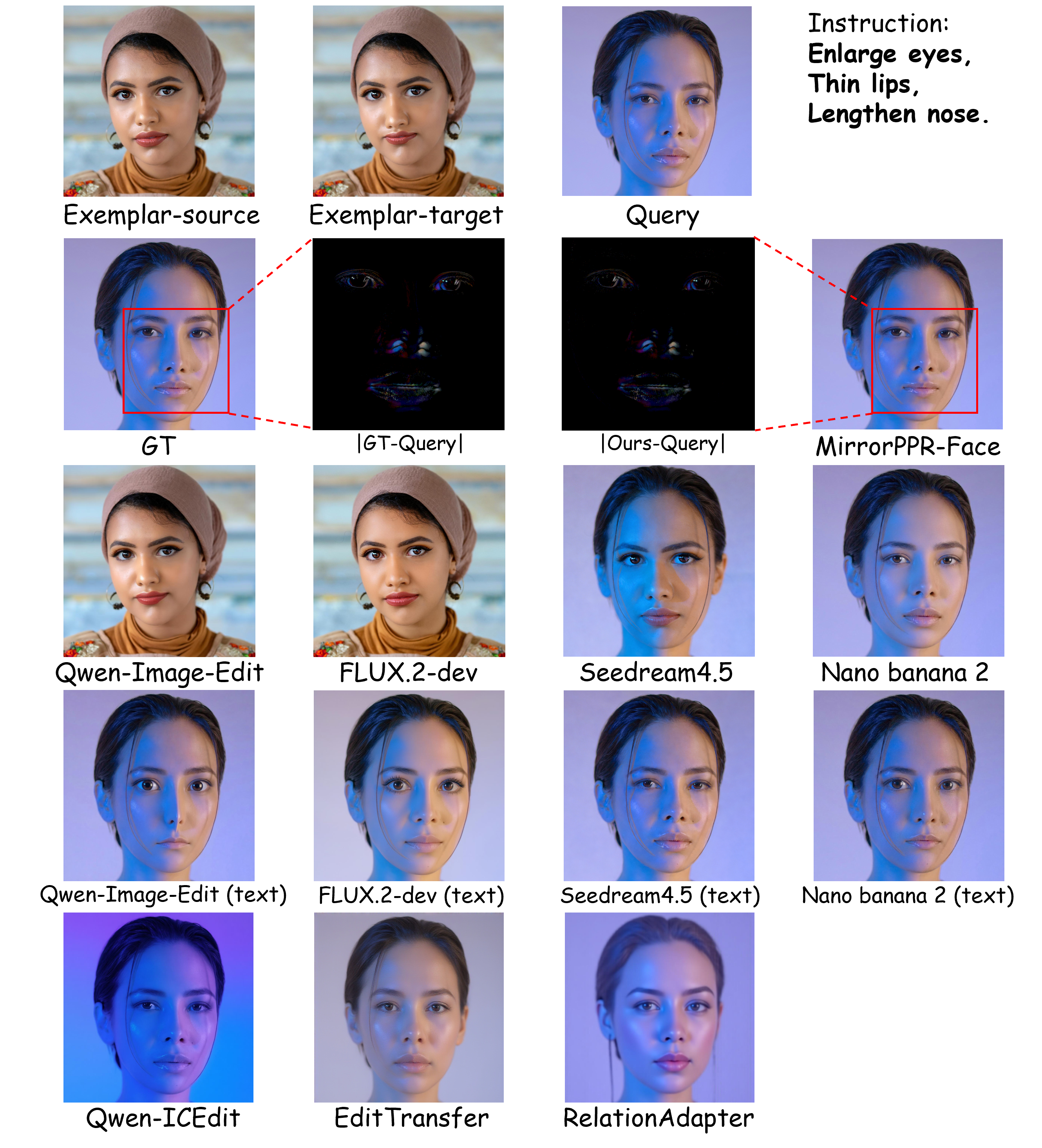}
  \caption{Qualitative comparisons between MirrorPPR and existing baselines on SimFace-100.}
  \label{fig:supple1}
\end{figure}

\begin{figure}[t]
  \centering
    \includegraphics[width=\textwidth]{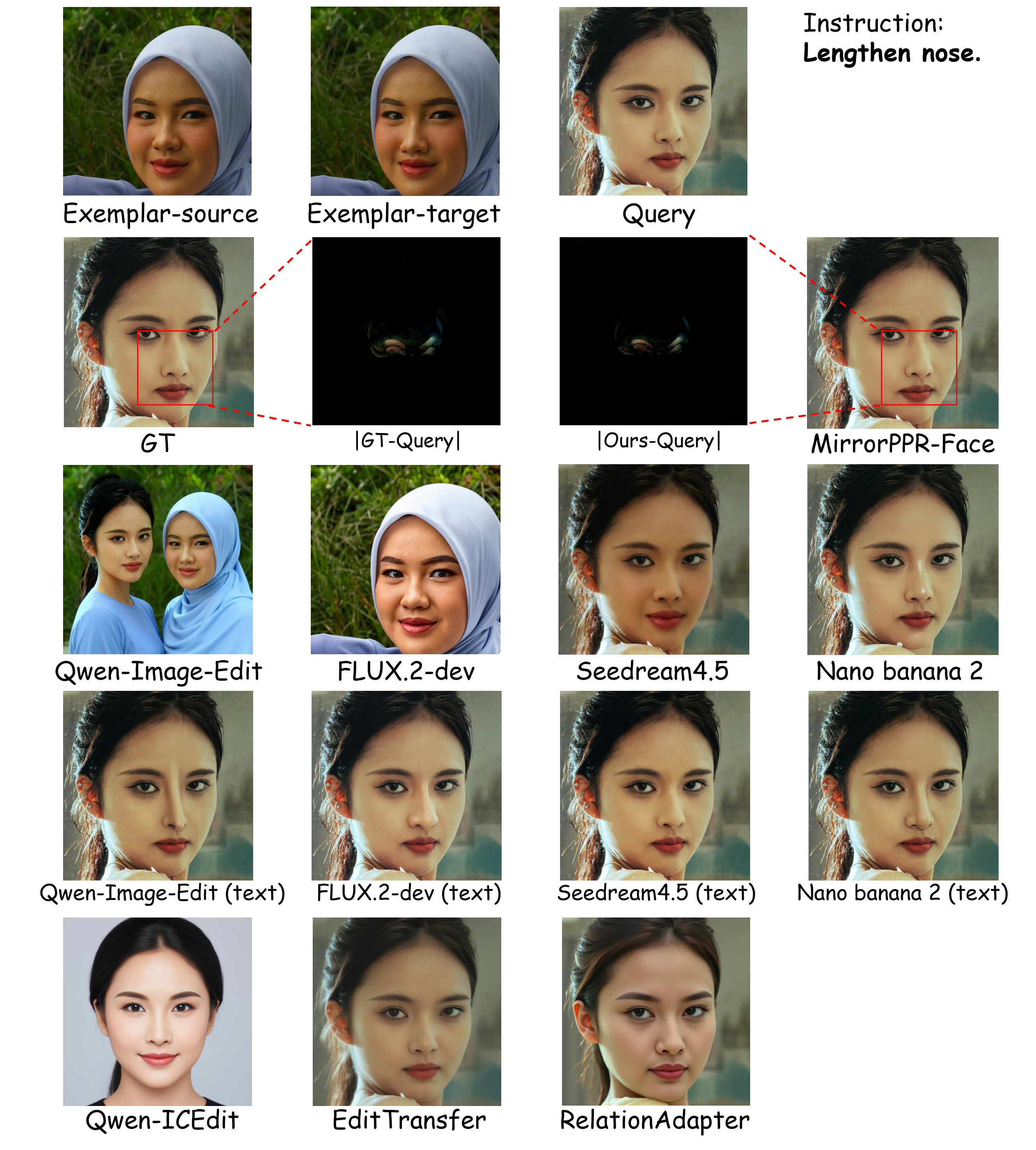}
  \caption{Qualitative comparisons between MirrorPPR and existing baselines on SimFace-100.}
  \label{fig:supple2}
\end{figure}

\begin{figure}[t]
  \centering
    \includegraphics[width=\textwidth]{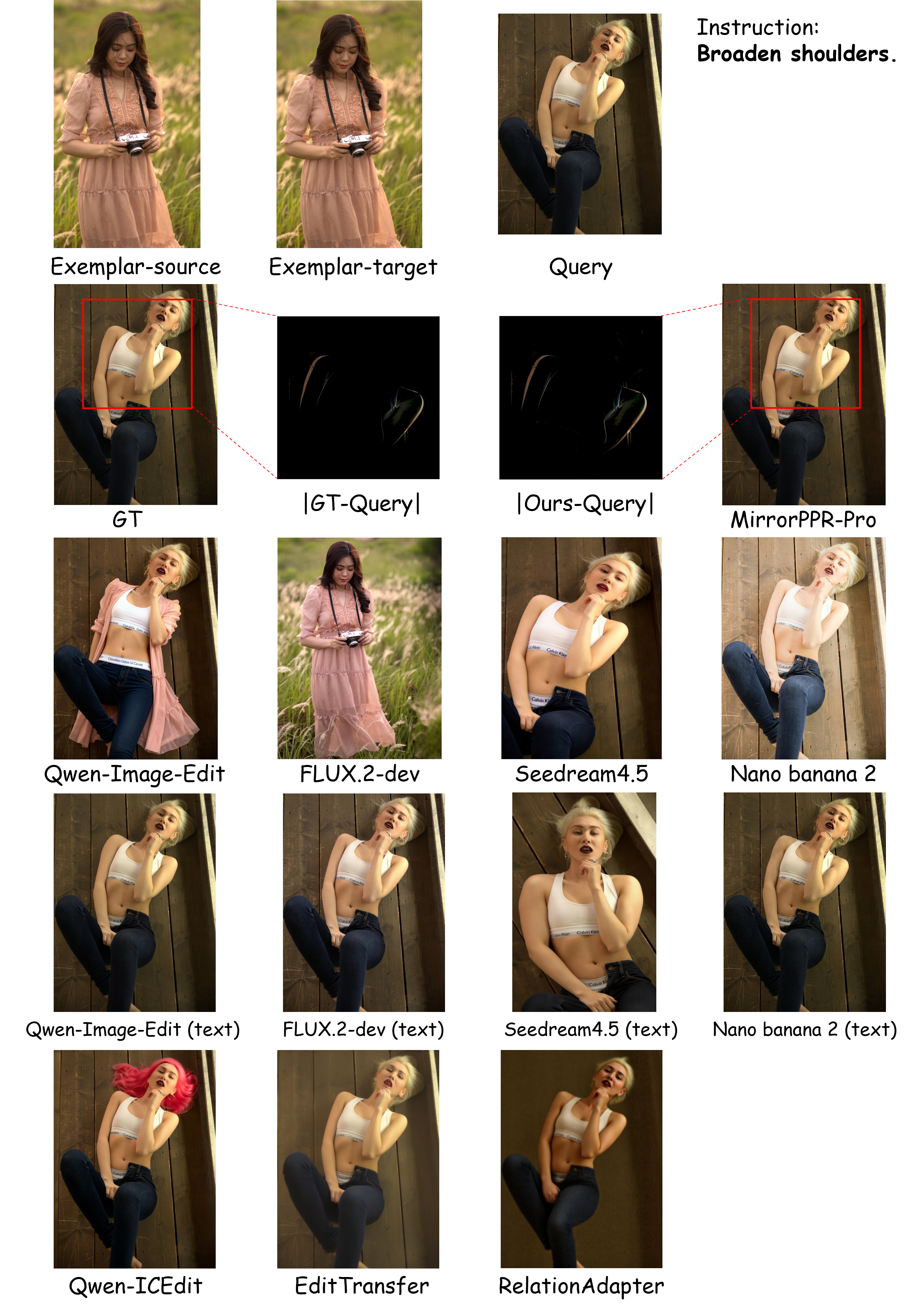}
  \caption{Qualitative comparisons between MirrorPPR and existing baselines on ProPortrait-500.}
  \label{fig:supple3}
\end{figure}

\begin{figure}[t]
  \centering
    \includegraphics[width=0.9\textwidth]{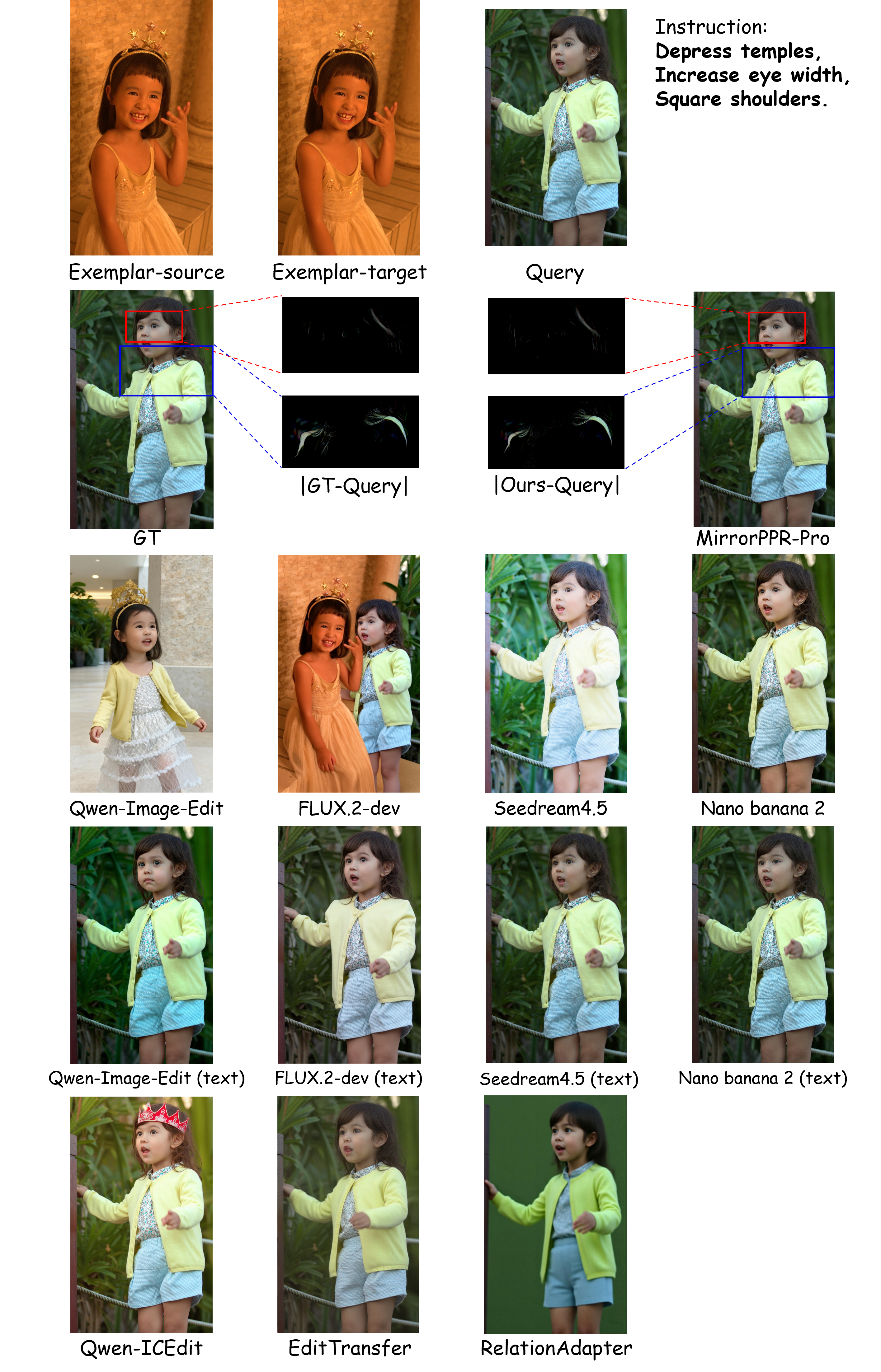}
  \caption{Qualitative comparisons between MirrorPPR and existing baselines on ProPortrait-500.}
  \label{fig:supple4}
\end{figure}

We provide additional qualitative comparisons between MirrorPPR and other baseline methods in Figure~\ref{fig:supple1} - Figure~\ref{fig:supple4}. These examples further demonstrate that our method can precisely capture and transfer the delicate retouching operations to the query image, whereas existing baselines suffer from various issues.

\clearpage

\end{document}